%% file: paper.tex
\documentclass[12pt]{l4dc2022}

\usepackage{enumitem}
\usepackage{tikz}
\usepackage[font=small,labelfont=bf]{caption}
\usetikzlibrary{arrows.meta}
\usepackage{booktabs}
\usepackage{wrapfig}

\newcommand{\methodname}{VSSF}
\newcommand{\fullmethod}{Variational State-Space Filters}

\newcommand{\linmethod}{L-VSSF}

\title[\linmethod]{Linear Variational State-Space Filtering}
\usepackage{times}
\author{%
 \Name{Daniel Pfrommer} \Email{dpfrom@seas.upenn.edu}%\\
 \AND
 \Name{Nikolai Matni} \Email{nmatni@seas.upenn.edu}%\\
}

\newcommand{\Meas}{{\mathcal X}}
\newcommand{\Data}{{\mathcal D}}
\newcommand{\Datadist}{{\mathcal D}}
\renewcommand{\L}{{\mathcal L}}
\newcommand{\E}{{\mathbb E}}
\newcommand{\DKL}{{\mathrm{D}_{\mathrm{KL}}}}
\newcommand{\R}{{\mathbb R}}
\newcommand{\N}{{\mathcal N}}
\newcommand{\modified}[1]{\overline{#1}}

\newcommand{\gener}{{ \theta }}
\newcommand{\infer}{{ \phi }}
\newcommand{\dyn}{{ \psi }}

\newcommand{\meas}{{x}}
\newcommand{\state}{{z}}
\newcommand{\inp}{{u}}

\newcommand\numberthis{\addtocounter{equation}{1}\tag{\theequation}}

\newtheorem{thm}{Theorem}[section]

\newtheorem{assum}[thm]{Assumption}

\DeclareMathOperator*{\argmax}{arg\,max}

\DeclareMathOperator*{\esssup}{ess\,sup}
\nocite{*}
\providecommand{\extended}{}

\ifdefined\extended
\jmlrvolume{}
\jmlryear{}
\jmlrproceedings{}{Extended Report}
\fi

\begin{document}

%tighten whitespace after captions a bit
\setlength{\belowcaptionskip}{-5pt}
\captionsetup{belowskip=-5pt}

%tighten display math a bit
\setlength{\abovedisplayskip}{4pt}
\setlength{\belowdisplayskip}{4pt}
\setlength{\abovedisplayshortskip}{4pt}
\setlength{\belowdisplayshortskip}{4pt}

\maketitle

\begin{abstract}
We introduce \fullmethod\ (\methodname), a new method for unsupervised learning, identification, and filtering of latent Markov state space models from raw pixels. We present a theoretically sound framework for latent state space inference under heterogeneous sensor configurations. The resulting model can integrate an arbitrary subset of the sensor measurements used during training, enabling the learning of semi-supervised state representations, thus enforcing that certain components of the learned latent state space to agree with interpretable measurements. From this framework we derive L-\methodname, an explicit instantiation of this model with linear latent dynamics and Gaussian distribution parameterizations. We experimentally demonstrate L-\methodname's ability to filter in latent space beyond the sequence length of the training dataset across several different test environments.
\end{abstract}

%\begin{keywords}%
%  List of keywords%
%\end{keywords}

\section{Introduction}

Representation learning is central to many difficult machine learning problems. Uncovering low dimensional embeddings of high dimensional data enables novel reasoning about generative processes, data compression \citep{theis2017lossy}, and probabilistic forecasting \citep{ibrahim2021variational}. Recent results in computer vision and natural language processing have demonstrated the effectiveness of large-scale generative models across difficult image \citep{yu2020an, zhang2020self} and language \citep{devlin2018bert} tasks.

Contemporary generative representation learning techniques based on Variational Autoencoders (VAEs) \citep{kingma2014autoencoding, rezende2014stochastic} and Generative Adversarial Networks (GANs) \citep{goodfellow2014generative} are capable of generating high fidelity and realistic looking still images by leveraging powerful latent representations \citep{oord2018neural}. Recent work using VAEs on video data and has shown the ability to predict future frames, \citep{babaeizadeh2017stochastic,denton2018stochastic} learn self-supervised representations of 3D structure \citep{lai2021video}, and enable compression ratios comparable to classical video codecs \citep{pessoa2020end}. In this paper we investigate the related problem of finding low dimensional embeddings suitable for control given pixel data and auxiliary low-dimensional (traditional) sensor measurements.

Applying deep representation learning techniques, specifically VAEs, to state space representations was  investigated by \citet{watter2015embed} and \citet{krishnan2015deep}, wherein a latent state embedding and locally linear latent state dynamics are jointly learned. More recent works such as Robust Controllable Embeddings \citep{banijamali2018robust}, PlaNet \citep{hafner2019learning}, Deep Variational Bayes Filters \citep{karl2017deep}, Dream to Control \citep{hafner2020dream}, Deep Kalman Smoother (DKS) \citep{krishnan2015deep}, and Online Variational Filtering (OVF) \citep{campbell2021online}, have extended these techniques to more challenging systems with nonlinear dynamics and partial observability. These methods have several key limitations. With the exception of DKS and OVF, these techniques either assume Markovian observation sequences \citep{watter2015embed, banijamali2018robust} or use amortized posterior approximations, where smoothing variational posteriors are only partially conditioned on future measurements \citep{karl2017deep, hafner2019learning, lee2019stochastic, hafner2020dream}. In the latter case, these approximations have been shown by \cite{bayer2021mind} to compromise the resulting the generative model, introducing a \emph{conditioning gap} suboptimality, while methods assuming Markovian observation sequences must batch several images together in order to infer otherwise hidden state, such as velocity. In either case, these structural limitations mean these methods are inherently not capable of modelling the proper filtering posteriors. 

In this work we apply variational inference techniques to learn a dynamics-consistent model for filtering in a low dimensional latent state space. Since we additionally have access to the full smoothing distribution, we can both model hidden system state and do not have to partially condition our inference distribution. Paired with suitably powerful neural network architectures, we demonstrate the ability of our method to infer dynamics-consistent filtering models using only image data, as well as integrate information from different sensor modalities simultaneously. Although we only discuss inference on state spaces with linear dynamics in this paper, the method we present is applicable to any system where closed-form filtering priors and inverse dynamics models can be computed.

\section{Related Work}
Prior work in deep stochastic video generation and prediction methods such as \citep{babaeizadeh2017stochastic, denton2018stochastic, lee2019stochastic}, differ subtly from the state space learning problem we consider in this paper. Video generation methods seek to model $p(x_T|x_{1:T-1})$, where $x_{1:T}$ is a sequence of $T$ video frames. By contrast, learned state space models are concerned with the inference distribution $p(z_{1:T}|x_{1:T}, u_{1:T-1})$ for some latent state $z_{1:T}$ and control input $u_{1:T-1}$. Whereas video generation models are free to pick an arbitrary latent state representation (e.g \cite{babaeizadeh2017stochastic} use one latent state $z$ across all frames $x_{1:T}$), state-space models are generally not free to do so.

Current deep variational state space learning methods can be broadly categorized into two classes: those that assume Markovian observation sequences \citep{watter2015embed, banijamali2018robust}, and those that allow for hidden state \citep{hafner2019learning, karl2017deep}. An observation sequence can generally be made Markovian by batching multiple image observations together, as done in \citep{watter2015embed,banijamali2018robust}. While this approach allows for models to only consider pairwise inference across temporally neighboring latent states $z_t$ and $z_{t+1}$, it requires the mapping from latent state $z_t$ to measurement $x_t$ to be deterministic in order to preserve the Markovian assumption.
%This has the advantage that the distribution $p(x_{1:T}|u_{1:T-1})$ can be factored  by $\prod_{t=1}^T p(x_t|x_{t-1}, u_{t-1})$ so models only need to consider inference for the pairwise $z_t, z_{t + 1}$. 
% However, while transitions $x_{t}$ to $x_{t + 1}$ may be stochastic, in order for the observation sequence to be Markovian the mapping from latent state $z_t$ to observation $x_t$ must be deterministic. Because each $x_t$ is in fact a batch of images, this simultaneously assumes that the transition dynamics for any hidden state is deterministic. 
This significant limitation has prompted work  \citep{krishnan2015deep, krishnan2016structured, hafner2019learning, bayer2021mind} into learning latent state spaces with both stochasticity and partial observability.

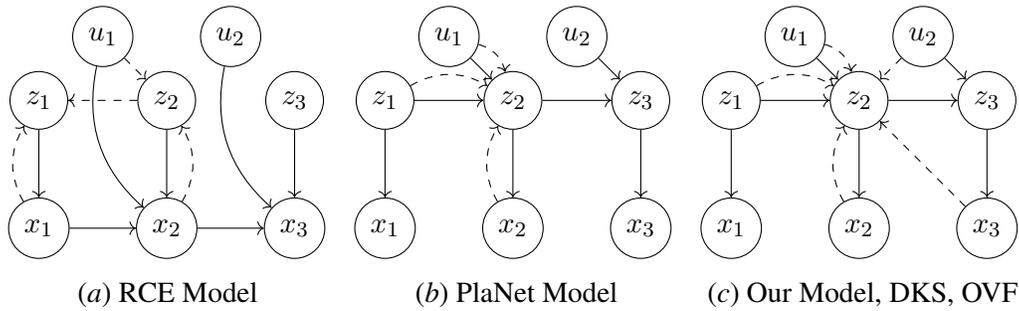
\begin{figure*}
\centering
\subfigure[RCE Model]{
\input{tikz/dep_rce.tikz}
}
\subfigure[PlaNet Model]{
\input{tikz/dep_planet.tikz}
}
\subfigure[Our Model, DKS, OVF]{
\input{tikz/dep_our.tikz}
}
\caption{Different latent model designs. Solid connections indicate the generative process, while dashed lines indicate the effective inference model for the future state $z_2$. Note (a) in \cite{banijamali2018robust} (RCE Model), both $z_1$ and $z_2$ are inferred using a deterministic reverse transition (b)  in \cite{hafner2019learning} (PlaNet),  there are no dependencies on future measurements $x_3$ or input $u_2$ when sampling $z_2$. Like \cite{campbell2021online} (OVF), our model (c) contains these dependencies but still enables access to the correct filtering distribution, unlike \cite{krishnan2016structured} (DKS)}
\label{fig:model}
\end{figure*}

Learning latent state spaces with partial observability requires the challenging task of learning the entire conditional latent state trajectory distribution $p(z_{1:T}|x_{1:T}, u_{1:T-1})$. PlaNet \citep{hafner2019learning} and related models \citep{hafner2020dream, lee2019stochastic} impose the factorization $p(z_{1:T}|x_{1:T}, u_{1:T-1}) = \prod_{t=1}^T p(z_t|z_{t-1}, x_{t}, u_{t-1})$. This effectively approximates the smoothing posterior $p(z_t|x_{1:T},u_{1:T-1})$ by the filtering posterior $p(z_t|x_{1:t},u_{1:t-1})$, incurring the conditioning gap suboptimality described in \cite{bayer2021mind}. The alternative approach taken by \cite{krishnan2016structured} in DKS is to factor the distribution as $p(z_{1:T}|x_{1:T}, u_{1:T-1}) = \prod_{t=1}^{T} p(z_{t}|z_{t-1}, u_{t-1:T}, x_{t:T})$. However, DKS has the significant drawback that estimating $z_t$ requires access to all future measurements $x_{t:T}$ as well as inputs $u_{t:T}$ and so does not provide access to a filtering distribution $p(z_t|x_{1:t})$.
In contrast to these aforementioned works, we use a decomposition for the full smoothing posterior of a partially observable state space model that allows for variational inference using the proper smoothing distribution at training time, but that also enables access to a filtering distribution suitable for use in real-time systems. The same smoothing decomposition is proposed in the concurrent work \cite{campbell2021online}, but we make the additional step of showing a closed form of the backwards smoothing distribution can be found, enabling simpler inference. The conceptual difference between these frameworks is illustrated in Fig. \ref{fig:model}.

\section{Problem Formulation}\label{sec:formulation}

Consider the stochastic dynamics given by the following Markov state space model
\begin{align} \label{eqn:gen}
    \state_1 \sim p_\gener(\state_1), \
    \state_{t + 1} \sim p_\dyn(\state_{t+1}|\state_t,\inp_t) %\label{eqn:dyn}
\end{align}
with state $\state_t$, input $\inp_t$, generative parameters $\gener$ and transition dynamics parameters $\dyn$.
 
At each time step $t$, the system emits a set of $k$ conditionally-independent observations $\Meas_t = \{\meas_t^{(j)} \}_{j=1}^k$ where
\begin{align} \label{eqn:meas}
    \meas_t^{(j)} &\sim p_\gener(\meas_t^{(j)}|\state_t).
\end{align}
The sensing modality can be different for each observation $\meas^{(j)}$, e.g., one observation could be an image obtained from a camera, and another a measurement obtained from an accelerometer. Further, in contrast to \cite{watter2015embed} and related models, we do not require the latent state $\state_t$ to be inferrable from $\Meas_t$, i.e., our state space model is only partially observable.

Our framework also allows for arbitrary subsets of training sensing modalities to be deployed at test time.  For example, suppose that training data is collected using both a motion capture system as well as images from a camera. Both the motion capture system and the camera data can be used during training, but the motion capture measurements can be omitted during testing.

Our goal is to learn both the parameters for the generative model $\theta$ as well as the system dynamics $\psi$. Formally, define $\Meas_{1:T} := \{\Meas_1, \Meas_2, \ldots, \Meas_{T}\}$ and $\inp_{1:T-1} := \{\inp_1, \inp_2, \ldots, \inp_{T-1}\}$ to be the $T$-length trajectories of measurements and control inputs, respectively. Given a set $\Data = \{ (\Meas^{(i)}_{1:T}, \inp^{(i)}_{1:T-1}) \}_{i=1}^n$ of $n$ independent input-observation trajectories of length $T$ sampled from the true data-generating distribution $p_\Datadist(\Data)$\footnote{We use the notation $p_\Datadist(X)$ to denote the distribution of $X$ as drawn from the true data generating distribution.}, we wish to find the maximum a posteriori estimate (MAP) for $\theta, \psi$ given by
\begin{align*}
    \hat{\gener}, \hat{\dyn} := \argmax_{\gener, \dyn} \: p(\gener, \dyn| \Data) = \argmax_{\gener, \dyn} \: p(\Data_\Meas | \dyn, \gener, \Data_\inp),
\end{align*}
under a uniform prior $p(\gener, \dyn)$ on the model parameters, and where have let $\Data_\Meas := \{\Meas^{(i)}_{1:T}\}_{i=1}^n$ and $\Data_\inp := \{\inp^{(i)}_{1:T-1}\}_{i=1}^n$.  Note that by independence of trajectories in the data set, the MAP estimate is equivalent to maximizing $\frac{1}{n}\sum_{i=1}^n\log p_{\gener, 
\dyn}(\Meas_{1:T}^{(i)}| \inp^{(i)}_{1:T-1}).$ 
To tackle this otherwise intractable optimization problem, we leverage variational inference techniques from \cite{kingma2014autoencoding, rezende2014stochastic} to introduce a variational approximation distribution $q_{\infer, \dyn}(\state_{1:T}|\Meas_{1:T}, \inp_{1:T-1})$ with inference parameters given by $\infer$. Following \cite{kingma2014autoencoding}, the resulting evidence-based lower bound (ELBO) is 
\begin{align*}
    \L(\infer, \gener, \dyn, \Meas_{1:T}, u_{1:T-1}) &= \log p_{\gener, \dyn}(\Meas_{1:T}| \inp_{1:T-1})  - G \\
    &= -\DKL(q_{\infer, \dyn}(\state_{1:T}|\Meas_{1:T}, \inp_{1:T-1}) || p_{\theta, \psi}(\state_{1:T}|\inp_{1:T-1}))) \\
    &\hspace{0.3cm} + \E_{q_{\infer, \dyn}(\state_{1:T}|\Meas_{1:T}, \inp_{1:T-1})}[\log p_{\gener, \dyn}(\Meas_{1:T}|\state_{1:T})], \numberthis \label{eqn:elbo}
\end{align*}
where $G = \DKL(q_{\infer, \dyn}(\state_{1:T}|\Meas_{1:T}, \inp_{1:T-1}) || p_{\gener, \dyn}(\state_{1:T}|\Meas_{1:T}, \inp_{1:T-1})))$ is the gap between the ELBO and the true log-likelihood of the generative conditional distribution.

Maximizing the ELBO is therefore equivalent to solving the original MAP parameter estimation problem under the following assumption.

\begin{assum} \label{assum:infer} The parametrization of the generating posterior $p_{\gener, \dyn}(\Meas_{1:T}|\state_{1:T}, \inp_{1:T-1})$ is sufficiently rich such that for optimal $\hat{\gener}, \hat{\dyn}$ and true data-generating distribution $p_\Datadist(\Meas_{1:T})$
\begin{align*}
    p_\Datadist(\Meas_{1:T}) = p_{\hat{\gener}, \hat{\dyn}}(\Meas_{1:T}),
\end{align*}
i.e., the true data-generating distribution can be feasibly modelled by our chosen parameterization. \\
Additionally, the optimal variational posterior $q_{\infer, \dyn}(\state_{1:T}|\Meas_{1:T}, \inp_{1:T-1})$ is sufficiently rich such that
\begin{align*}
    \esssup_{(\Meas, \inp) \sim p_\Datadist} \DKL(q_{\hat{\infer}, \hat{\dyn}}(\state_{1:T}|\Meas_{1:T}, \inp_{1:T-1}) || p_{\hat{\gener}, \hat{\dyn}}(\state_{1:T}|\Meas_{1:T}, \inp_{1:T-1}))) = 0,
\end{align*}
i.e., the gap between the lower bound and the true model log-likelihood can be driven to zero for the MAP estimate of $\gener, \dyn$.
\end{assum}

\begin{sloppypar}
These two conditions imply that (a) the inferred latent state distribution prior $q_{\hat{\infer}}(\state_t) = \int q_{\hat{\infer}}(\state_t|\Meas_t)p_\Datadist(\Meas_t)d\Meas_t$ is equivalent to $p_{\hat{\gener}}(\state_t)$ for all $t$, and (b) because the evidence lower bound is tight for the MAP parameter estimate, maximizing the lower bound is equivalent to solving the MAP optimization problem. While this is a very restrictive assumption, this is only used to show the correctness of subsequent problem simplifications in Section \ref{sec:obs}. Note that in practice it is sufficient for these conditions to be approximately satisfied in order to obtain good performing models. We leave the problem of quantifying the effect of approximately satisfying Assumption \ref{assum:infer} on the resulting model degradation to future work.
\end{sloppypar}

\subsection{Smoothing Inference}
We now propose a factorization of the full smoothing posterior $q_{\infer, \dyn}(\state_{1:T}|\Meas_{1:T}, \inp_{1:T-1})$ in terms of the observation inference model $q_\infer(\state_t|\Meas_t^{(j)})$, state prior $p_\gener(\state_t)$, and transition model $p_\dyn(\state_{t + 1}|\state_{t}, \inp_t)$, by leveraging the joint distribution smoothing decomposition used in \cite{campbell2021online, briers2010smoothing}. Rather than adopting a forward factorization of the smoothing posterior $q_{\infer, \dyn}(\state_{1:T}|\Meas_{1:T}, \inp_{1:T-1})$, wherein the current latent state $z_t$ is dependent on the previous latent state $z_{t - 1}$, we instead model the \emph{backwards in time} dependency of $z_t$ on $z_{t + 1}$ using the decomposition  %$q_{\infer, \dyn}(\state_t | \state_{t + 1}, \Meas_{1:t}, u_{1:t})$
\begin{align}
    q_{\infer, \dyn}(\state_t | \state_{t + 1}, \Meas_{1:t}, \inp_{1:t}) &\propto q_{\dyn}(\state_{t + 1} | \state_t, \inp_t)q_{\infer, \dyn}(\state_t|\Meas_{1:t}, \inp_{1:t-1}). \label{eqn:factor}
\intertext{Therefore given a latent state $z_{t + 1}$, the distribution over the previous latent state $z_t$ can be found by: (i) computing the filtering posterior $q_{\infer, \dyn}(z_t |\Meas_{1:t}, u_{1:t-1})$, (ii) finding a closed form for the transition distribution $q_{\dyn}(\state_{t + 1} | \state_t, \inp_t) = p_{\dyn}(\state_{t + 1} | \state_t, \inp_t)$ as a function $\state_t$, and (iii) taking the product of the densities.  In particular, for linear dynamics and Gaussian distribution setting we consider in \S\ref{sec:lvssm}, all of these steps admit closed-form expressions. Furthermore because the full smoothing posterior factors as the product}
    q_{\infer, \dyn}(\state_{1:T}|\Meas_{1:T}, \inp_{1:T-1}) &= q_{\infer, \dyn}(\state_{T}|\Meas_{1:T}, \inp_{1:T-1})\prod_{t=1}^{T-1} q_{\infer, \dyn}(\state_t | \state_{t + 1}, \Meas_{1:t}, \inp_{1:t}), \label{eqn:bkw}
\end{align}
the samples and associated likelihoods for $q_{\infer, \dyn}(\state_{1:T}|\Meas_{1:T}, \inp_{1:T-1})$ can be efficiently computed recursively by first sampling the final latent state $\state_T$ from the marginal $q_{\infer, \dyn}(\state_{T}|\Meas_{1:T}, \inp_{1:T-1})$, and
then recursively sampling from the corresponding marginals $q(\state_t | \state_{t + 1}, \cdot)$ for $t={T-1}, {T-2}, \ldots 1$.

Sampling from $q_{\infer, \dyn}(z_{1:T}|\Meas_{1:T},u_{1:T-1})$ is therefore a two-pass algorithm. We first perform a forward pass to compute the filtering posteriors $q_{\infer, \dyn}(\state_t|\Meas_{1:t}, u_{1:t-1})$ using the standard propagation and update recursions \citep{tanizaki1996nonlinear}
\begin{align}
    q_{\infer, \dyn}(\state_{t}|\Meas_{1:t-1}, \inp_{1:t-1}) &= \int q_\dyn(\state_{t}|\state_{t -1}, \inp_{t-1})q_\infer(\state_{t-1}|\Meas_{1:t-1}, \inp_{1:t-2})d\state_{t-1}, \\
    q_{\infer, \dyn}(\state_{t}|\Meas_{1:t}, \inp_{1:t-1}) &\propto q_{\infer, \dyn}(\state_t, | \Meas_{1:t-1}, \inp_{1:t-1})\prod_{j=1}^k \frac{q_{\infer}(z_t|\meas_{t}^{(j)})}{q_{\infer, \dyn}(z_t)}. \label{eqn:update}
\end{align}
Then we perform a backwards pass to sample a (backwards) trajectory $\state_{T:1}$ using equations \eqref{eqn:factor} and \eqref{eqn:bkw}. 
\ifdefined\extended
This two-pass approach to sampling $\state_{1:T}$ is visualized in Figure \ref{fig:flow}.
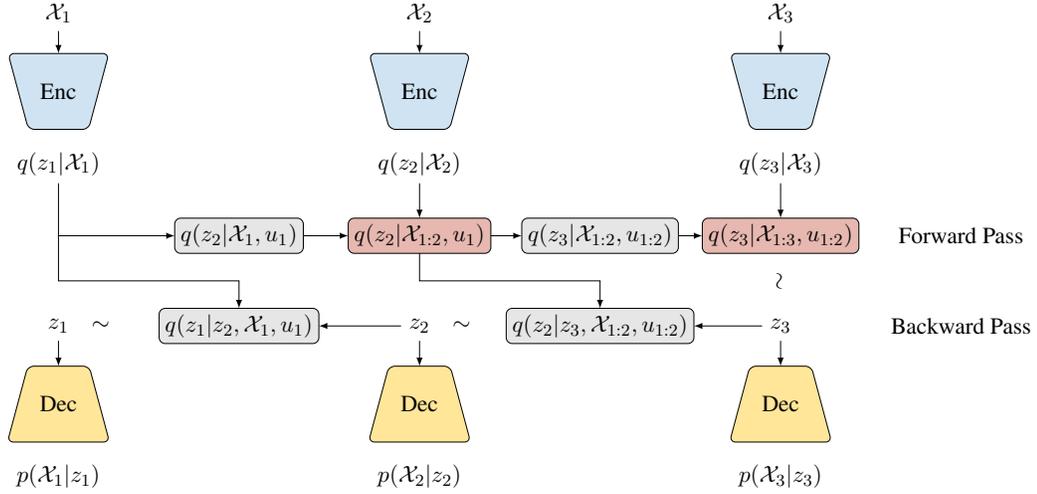
\begin{figure}
    \centering
    \scalebox{0.8}{
    \input{tikz/flow.tikz}
    }
    \caption{Visualization of the two-pass sampling approach for the smoothing posterior $q(\state_{1:T}|\Meas_{1:T},\inp_{1:T-1})$ and corresponding reconstruction $p(\Meas_t|\state_t)$. The red boxes show the observation-updated distributions and the grey boxes show the filtering-forwards/smoothing-backwards propagated distributions.}
    \label{fig:flow}
\end{figure}
\fi
Modelling the distribution $q_{\infer, \dyn}(z_t)$ needed for the marginal posterior $q_{\infer, \dyn}(\state_{t}|\Meas_{1:t}, \inp_{1:t-1})$ is generally intractable, but Assumption \ref{assum:infer} allows setting $q_{\infer, \dyn}(\state_t) = p_{\theta}(\state_t)$ without affecting the correctness of the resulting model given proper parameterizations (see \S\ref{append:correctness}, as well as the conditions required for proper inference of per-observation $q_{\infer}(z_t|\meas_{t}^{(j)})$ discussed in \S\ref{append:correct_filter}). 

With this sampling procedure, we can compute a Monte-carlo approximation for the KL term
$\DKL(q_{\infer, \dyn}(\state_{1:T}|\Meas_{1:T}, \inp_{1:T-1}) || p_{\theta, \psi}(\state_{1:T}|\inp_{1:T-1}))$ of the ELBO \eqref{eqn:elbo}, as described in \cite{kingma2014autoencoding}, as the prior log-likelihood factors as 
$$\log p_{\theta, \psi}(\state_{1:T}|\inp_{1:T-1}) = \log p_{\theta}(\state_1) + \sum_{t=1}^{T-1}\log p_\dyn(\state_{t + 1}|\state_{t}, \inp_t),$$
due to the Markov structure of the latent state. Similarly we can approximate the reconstruction term $\E_{q_{\infer, \dyn}(\state_{1:T}|\Meas_{1:T}, \inp_{1:T-1})}[\log p_{\gener, \dyn}(\Meas_{1:T}|\state_{1:T})]$ of the ELBO \eqref{eqn:elbo} by leveraging conditional independence of the measurements $\meas_t^{(j)}$ given the latent state $z_t$:
$$\log p_{\gener, \dyn}(\Meas_{1:T}|\state_{1:T}) = \sum_{t=1}^T \sum_{j=1}^k \log p_{\gener, \dyn}(\meas_t^{(j)}|\state_t).$$

In the next section, we show that under suitable parameterizations of the dynamics $p_\dyn(\state_{t + 1}|\state_t, \inp_t)$, generative model $p_\theta(\Meas_t|\state_t)$ and inference models $q_\infer(\state_t|\Meas_t)$ we can efficiently maximize the ELBO \eqref{eqn:elbo}.  
\ifdefined\extended
Detailed derivations of the results presented in this section con be found in Appendix \ref{appendix:model}.
\else
Detailed derivations of the results presented in this section can be found in the extended version \cite{extended}.
\fi

\section{Variational State Space Models with Linear Dynamics} \label{sec:lvssm}
We consider an instantiation of the above framework with linear dynamics and Gaussian distributions. The general approach presented above is independent of these choices, and any combination of discrete or continuous latent variables with different distribution parameterizations can in principle be used for the latent state representation.

As a motivating example we consider the object tracking problem discussed in Section \ref{sec:formulation} where the latent state may be partially measured through a secondary sensor such as an accelorometer. In order to facilitate representation learning with partially known latent state, we will introduce two distinct observation models: a nonlinear model suitable for image data, as well as a linear model that can directly observe (subsets of) the latent state.

Consider a Markov state space model as described in Section \ref{sec:formulation}, where the latent state $\state_t \in \R^m$ and the system evolves according to linear dynamics of the form
\begin{align*}
    \state_1 \sim \N(0, \Sigma_z), \
    \state_{t + 1} = A\state_t + B\inp_t + w_t, \
    w_t &\overset{\scriptsize{i.i.d.}}{\sim} \N(0, \Sigma_w),
\end{align*}
where the prior covariance $\Sigma_z$ is fixed. The state space parameters $(A, B, \Sigma_w)$ are treated either as known parameters or consolidated into the unknown parameters $\dyn$, depending on the problem setup.  We parameterize the measurement model $q_\infer(z_t|\meas_{t}^{(j)})$ as a Gaussian such that under Assumption \ref{assum:infer} we have that
\begin{align}
\frac{q_{\infer}(\state_t|\meas_{t}^{(j)})}{q_{\infer, \dyn}(\state_t)} \approx \frac{q_{\infer}(\state_t|\meas_{t}^{(j)})}{p_{\theta}(\state_t)} &\propto \N(h_t^{(j)}, H_t^{(j)}), \label{eqn:mn_meas}
\end{align}
where $h_t^{(j)}$ and $H_t^{(j)}$ are computed from $\meas_t^{(j)}$ according to the model chosen for $p_\theta(\meas_t^{(j)}|\state_t)$. We will discuss parameterizations of $h_t^{(j)}$ and $H_t^{(j)}$ for linear and nonlinear measurement models in \S\ref{sec:nonlinear_obs} and \S\ref{sec:observation}.

Combining equation \ref{eqn:mn_meas} with the linear dynamics model above, we find closed-form expressions for the mean and covariances of the resulting Gaussian filtering prior $q_{\infer, \dyn}(\state_t|\Meas_{1:t-1}, u_{1:t-1})$ and filtering posterior $q_{\infer, \dyn}(\state_t|\Meas_{1:t}, u_{1:t-1})$. Letting the filtering prior $q_{\infer, \dyn}(\state_t|\Meas_{1:t-1}, u_{1:t-1}) \sim \N(p_{t|t-1}, P_{t|t-1})$ and the filtering posterior $q_{\infer, \dyn}(\state_t|\Meas_{1:t}, u_{1:t-1}) \sim \N(p_{t|t}, P_{t|t})$, the priors can be computed recursively using the standard Kalman filter propagation equations \citep{terejanu2009discreteKF}:
\begin{align*}
    P_{t|t-1} = AP_{t-1|t-1}A^\top + \Sigma_\state, \quad
    p_{t|t-1} = Ap_{t-1|t-1} + B\inp_{t}.
\end{align*}
Similarly, the posterior $\N(p_{t|t}, P_{t|t})$ can be computed recursively in terms of the information matrix $P_{t|t}^{-1}$ and information vector $P_{t|t}^{-1}p_{t|t}$ using the Information Filter update equations \citep{terejanu2009discreteKF}:
\begin{align*}
    P_{t|t}^{-1} = P_{t|t-1}^{-1} + \sum_{j=1}^k (H_t^{(j)})^{-1}, \quad P_{t|t}^{-1}p_{t|t} = P_{t-1|t-1}^{-1}p_{t-1|t-1} + \sum_{j=1}^k (H_t^{(j)})^{-1}h_t^{(j)}.
\end{align*}
The use of the Information filter update as opposed to the standard state-space based update allows for easy simultaneous fusion of information from multiple observations. The reverse smoothing distribution $q_{\infer, \dyn}(\state_t|\state_{t + 1},\Meas_{1:T}, \inp_{1:T-1}) = \N(\ell_t, L_t)$ can likewise be computed in terms of the filtering posterior $\N(P_{t|t}, p_{t|t})$ and next state $\state_{t + 1}$
\begin{align*}
    L_t^{-1} = P_{t|t}^{-1} + A^\top \Sigma_{w}^{-1}A, \ L_t^{-1}\ell_t = A^\top\Sigma_{w}^{-1}(P_{t|t}^{-1} - B \inp_t) + P_{t|t}^{-1}p_{t|t}.
\end{align*}

We will now discuss different observation models for $\frac{q_{\infer}(\state_t|\meas_{t}^{(j)})}{p_{\theta}(\state_t)} \propto \N(h_t^{(j)}, H_t^{(j)})$ and $p_\gener(\meas_t^{(j)}|\state_t)$ that can be used within this framework.

\subsection{Observation Models}\label{sec:obs}

We make the following simplifying assumption.

\begin{assum} \label{assum:invariant_prior} The true underlying prior distribution $p_\Datadist(\state_t)$, state-conditional observation distribution $p_\Datadist(\Meas_t | \state_t)$, and state inference distributions $p_\Datadist(\state_t | \Meas_t)$ are time-invariant 
\end{assum}

Using time-varying prior $p_\theta(z_t)$ would require the state inference distribution $q_\infer(\state_t|\Meas_t)$ to also be time-varying. This is not only generally undesirable and computationally infeasible for large neural networks, but the resulting model would not be usable for sequences over a horizon longer than $T$, the horizon of the training data trajectories. Assumption \ref{assum:invariant_prior} is satisfied, for example, if the initial state $p_\Datadist(\state_1)$ is equal to the steady-state distribution of the closed-loop system from which the data is generated and the observation model $p_\Datadist(\Meas_t|\state_t)$ is identical for all $t$.

\subsubsection{Nonlinear Observation Model}\label{sec:nonlinear_obs}
To handle nonlinear and high-dimensional measurements $\meas_t^{(j)} \in \R^p$ such as raw pixel data, we introduce a nonlinear observation model. In this case we parameterize the generative model $p_\theta(\meas_t^{(j)}|\state_t)$ for nonlinear sensing modality $j$ as $p_\theta(\meas_t^{(j)}|\state_t) = \N(\nu(\meas_t^{(j)}), \Sigma_\meas)$
for neural network-parameterized mean $\nu(\meas_t^{(j)})$ and fixed covariance matrix $\Sigma_\meas$,
\footnote{Current state of the art variational autoencoders such as VQ-VAE \citep{oord2018neural} and VDVAE \citep{child2020very} use discrete logistic distribution mixtures parameterizations for $p_\theta(\meas_t^{(j)}|\state_t)$. We use a diagonal normal distribution in our experiments since it is straightforward to implement and corresponds directly to a mean-square-error loss term in the resulting ELBO while still producing good results. We leave the problem of tuning this model for better performance to future work.}
and with corresponding inference model $q_\infer(z_t|\meas_t^{(j)}) = \N(r_h(\meas_t^{(j)}), r_H(\meas_t^{(j)}))$,
for neural network-parameterized mean function $r_h$ and positive definite neural network-parameterized coveriance matrix $r_H$. In Section \ref{sec:experiment} we discuss neural-network parameterizations of this model suitable for inference from image data, as well as how to parameterize $r_H$ in order to enforce this constraint.

\subsubsection{Linear Observation Model}
\label{sec:observation}
\label{sec:linear_obs}

Consider the standard linear measurement model where
\begin{align*}
    \meas_t^{(j)} = C^{(j)}\state_t + w_\meas,\  
    w_\meas \overset{\scriptsize{i.i.d.}}{\sim} \N(0, \Sigma_\meas),
\end{align*}
where $C^{(j)}\in \R^{p \times n_z}$, and $p<n_z$, i.e., a noisy observation from a low-dimensional subspace of the latent-state is measured. 
In this case a closed form for both $p_\theta(\meas_t^{(j)}|\state_t)$ and $p_\theta(\state_t|\meas_t^{(j)})$ can be found, so the variational approximation $q_\infer(\state_t|\meas_t^{(j)})$ is not needed, and the posterior $p_\theta(\state_t|\meas_t^{(j)})$ can be used directly:
\begin{align*}
p_\theta(\meas_t^{(j)}|\state_t) \sim \N(C^{(j)}\state_t, \Sigma_x), \
p_\theta(\state_t|\meas_t^{(j)}) \sim \N(\mu, \Sigma),
\end{align*}
where $\Sigma^{-1} = C^\top\Sigma_x^{-1}C + \Sigma_z^{-1}$ and $\Sigma^{-1}\mu = C^\top \Sigma_x^{-1}x_t^{(j)}$.

Note that under Assumption \ref{assum:infer} it follows that $(H^{(j)})^{-1} = C^\top \Sigma_\meas^{-1}C$ and $(H^{(j)})^{-1}h^{(j)} = C^\top \Sigma_\meas^{-1}\meas_t^{(j)}$, which correspond to the measurement information matrix/vector of a standard information filter \citep{terejanu2009discreteKF}.

By fixing the appropriate sparse $C^{(j)} \in \R^{s \times m}$ and combining the linear observation model with nonlinear observations, situations where the state is partially known or where a subset of the dataset is labelled can be cleanly handled.\footnote{If $C^{(j)}$ is fixed, the prior $p_\theta(z)$ should be chosen to be consistent with the true distribution $p_\Datadist(C^{(j)}z)$.} We demonstrate this ability in the synthetic experiments presented in Section \ref{sec:experiment}. 

% Acknowledgments---Will not appear in anonymized version
%\acks{We thank a bunch of people.}

\section{Experiments}\label{sec:experiment}
\begin{figure}
    \centering
    \includegraphics[height=180px]{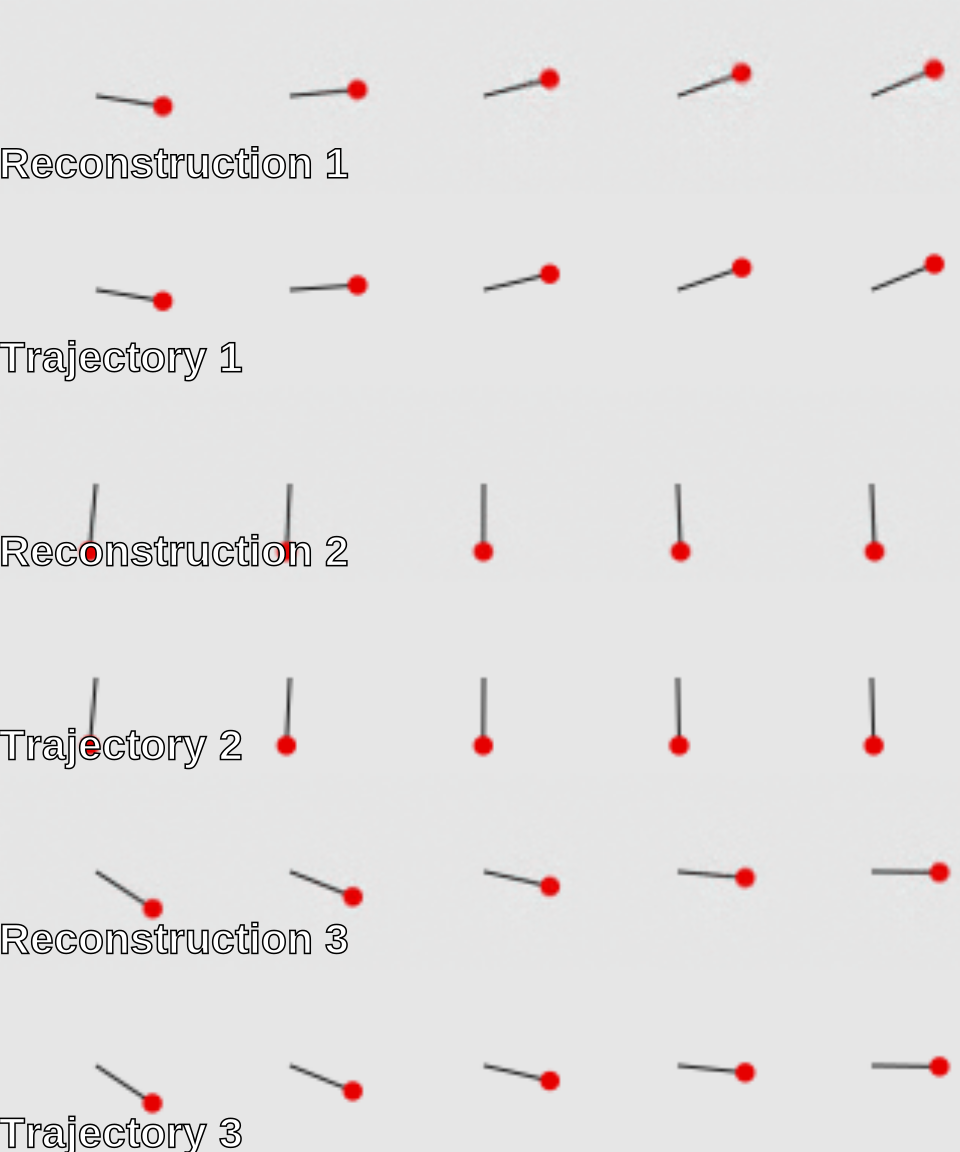}
    \includegraphics[height=180px]{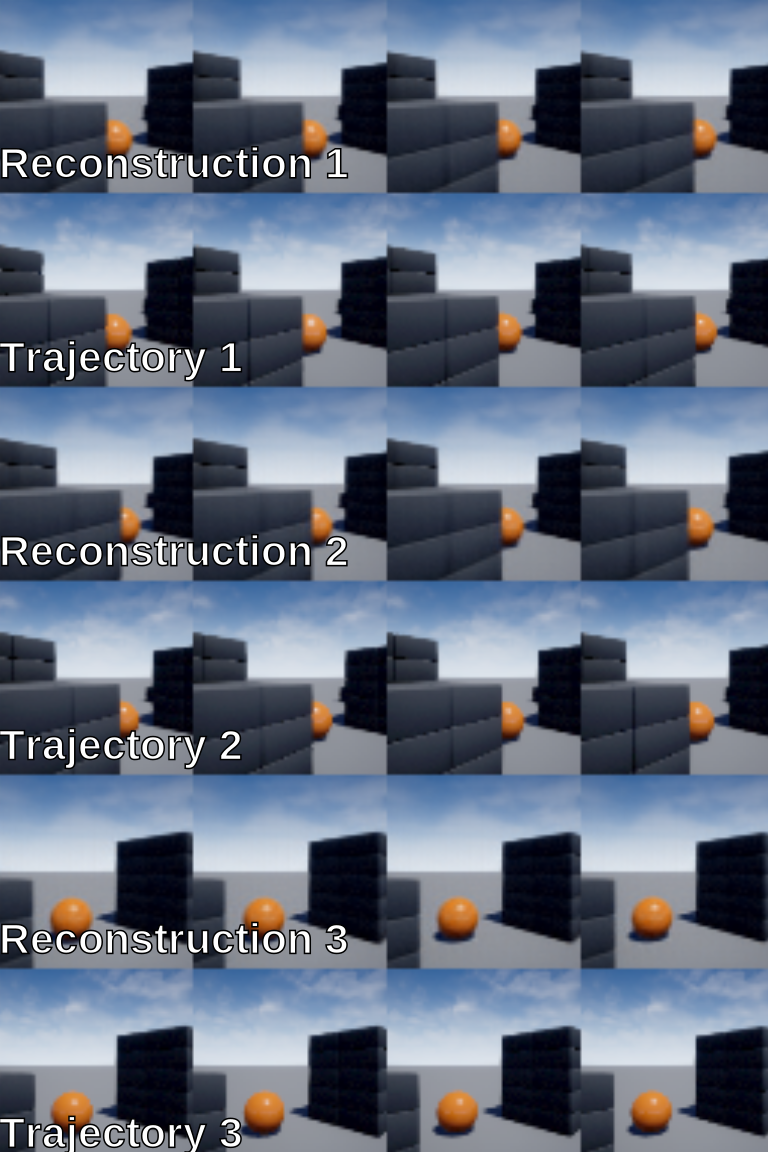}
    \includegraphics[height=180px]{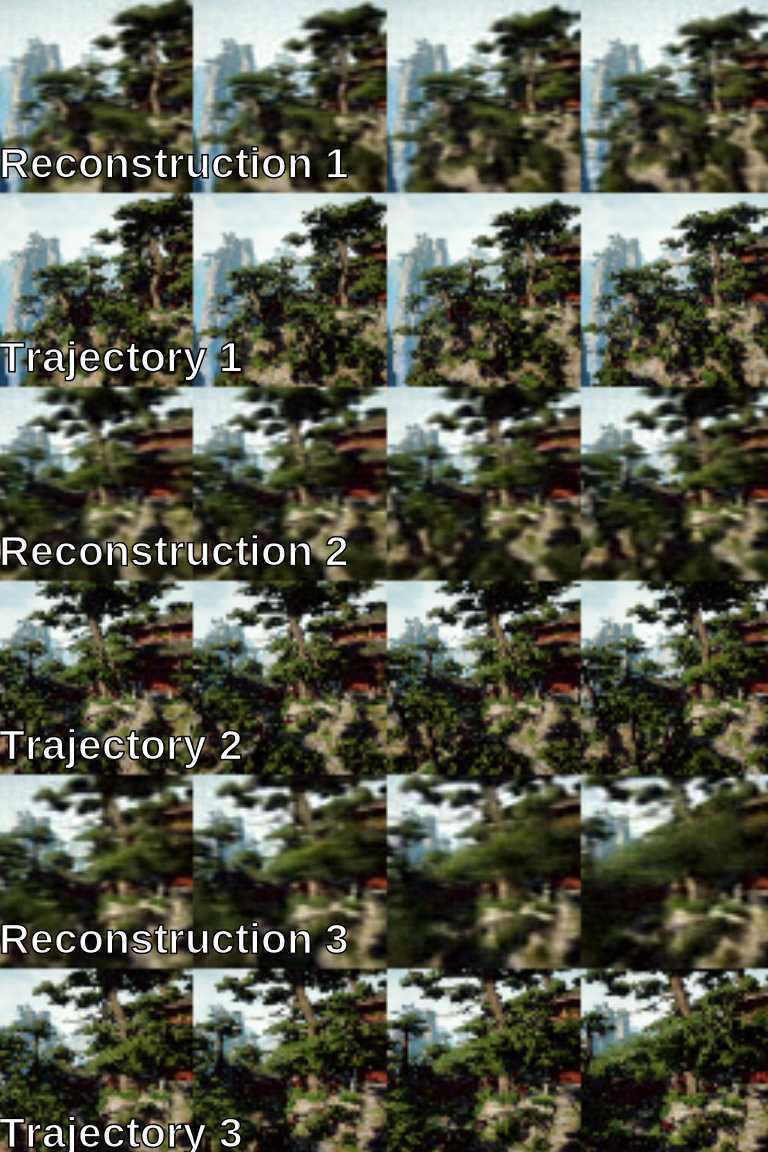}
    \caption{Reconstructions for three test trajectories from each of the pendulum, Blocks, and Zhangjiajie environments under fixed $\dyn$. We speculate that the reconstructions for the Zhangjiajie environment are worse due to the feature-rich environment and dynamic fog effects present. The complex cloud patterns in the Blocks environment demonstrate a similar blurring effect.}
    \label{fig:reconst}
\end{figure}

We performed experiments in the following environments.
\begin{itemize}
\item \textbf{Pendulum Environment:}
For the pendulum environment depicted in Figure \ref{fig:reconst}, our dataset consists of $10000$ sequences of length $T = 5$. We used a $3.14\tanh(\theta/3.14)$ nonlinearity before rendering the final pendulum images in order to constrain the visual angle of the pendulum to the $(-\pi, \pi)$ range, even if the magnitude of $\theta$ sporadically exceeds $\pi$. 
\item \textbf{Blocks Environment:}
For the second environment we moved a camera in the Unreal-Engine-based Airsim simulator \citep{airsim2017fsr} with double-integrator dynamics in $x,y$ and fixed height $z$ as well as fixed camera heading. The goal in this problem is to learn a mapping from camera images to $(x,y)$-coordinates, a task analogous, but not equivalent to, visual-inertial-odometry (VIO). We used a dataset of size 10000 and trajectory length $T = 4$ in the "Blocks" Airsim environment.
\item \textbf{Zhangjiajie Environment:} We additionally created a third dataset with the same double-integrator dynamics as the Blocks dataset, set in the "Zhangjiajie" Airsim environment. The Blocks and Zhangjiajie environments are both visualized in Fig \ref{fig:reconst}. The Zhangjiajie environment is much more complex than the pendulum or block environments and we perform correspondingly worse on this dataset.
\end{itemize}

We primarily consider the case where the latent state dynamics parameters $\dyn$ are known, and seek to learn the model parameters $\gener, \infer$. We have found that simultaneously optimizing over $\dyn, \gener, \infer$ is difficult due to the tendency for the state transition matrix $A$ to degenerate and cause posterior collapse. We hypothesize that prior work \citep{karl2017deep,krishnan2016structured,krishnan2015deep, hafner2019learning, hafner2020dream} does not suffer from this issue since the estimates of the initial state $\state_1$ are made independently of the dynamics parameters, whereas we model $z_1$ as dependent on $z_{2:T}$ and the system dynamics. In practice the collapse of $q_\infer(z_t)$ can be avoided through the introduction of a direct linear state observation model.

\ifdefined\extended
\else
\begin{wrapfigure}{r}{.48\textwidth}
    \centering
\subfigure[][l]{
    \includegraphics[width=95px]{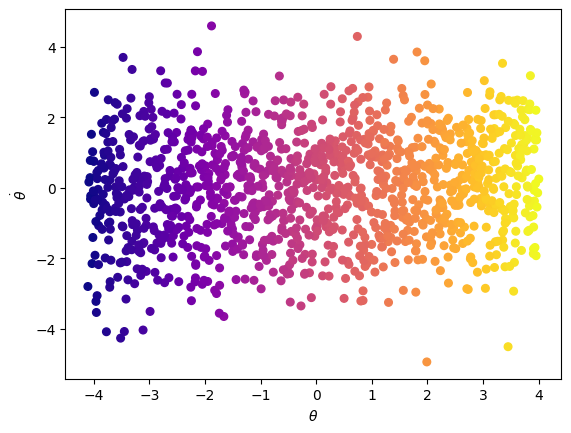}
}~\subfigure[][l]{
    \includegraphics[width=95px]{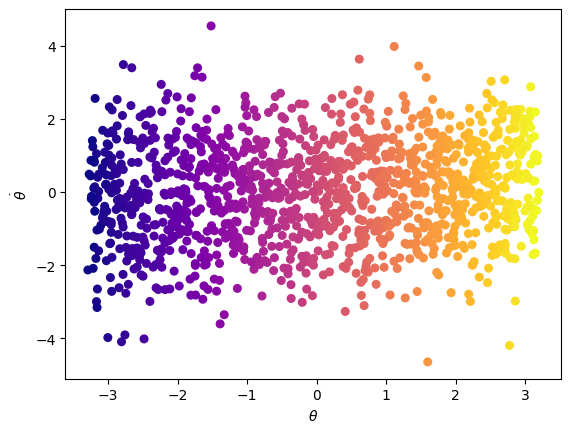}
} \\
\subfigure[][l]{
    \includegraphics[width=95px]{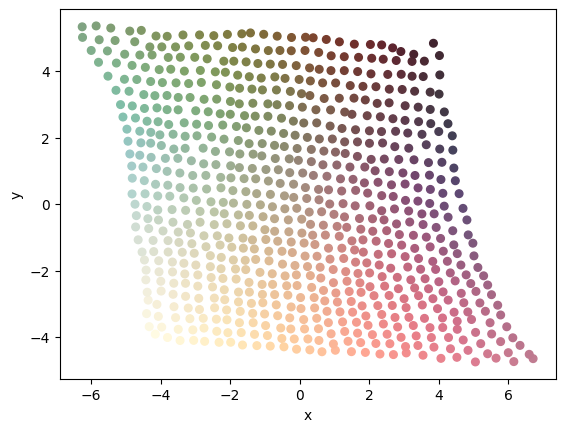}
}~\subfigure[][l]{
    \includegraphics[width=95px]{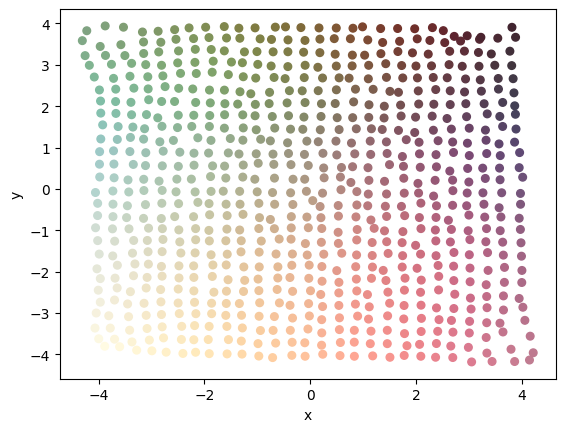}
}
    \caption{Visualizations of the latent space for (a) pendulum, using only image data, (b) pendulum, partially supervised, (c) $x,y$ of the Block environment from images, (d) $x, y$ of the Block environment, partially supervised.}
    \label{fig:latent}
\end{wrapfigure}
\fi

\begin{figure}
    \centering
    \subfigure[\small Image-based inference using pixel data only.]{
    \includegraphics[height=150px]{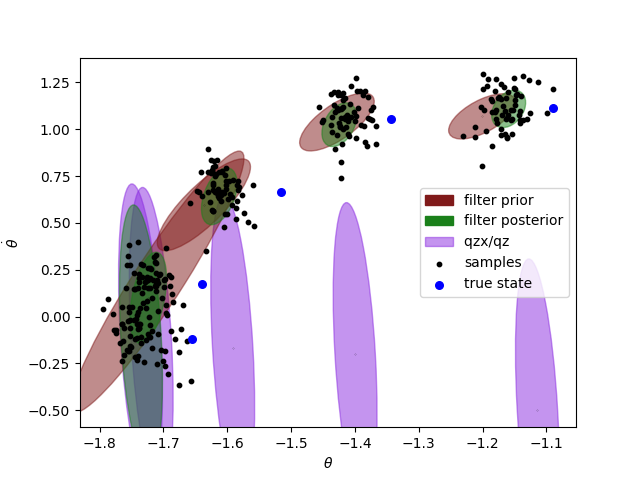}
    }
    \subfigure[\small Image-based inference with direct $\theta$ observations at training-time.]{
    \includegraphics[height=150px]{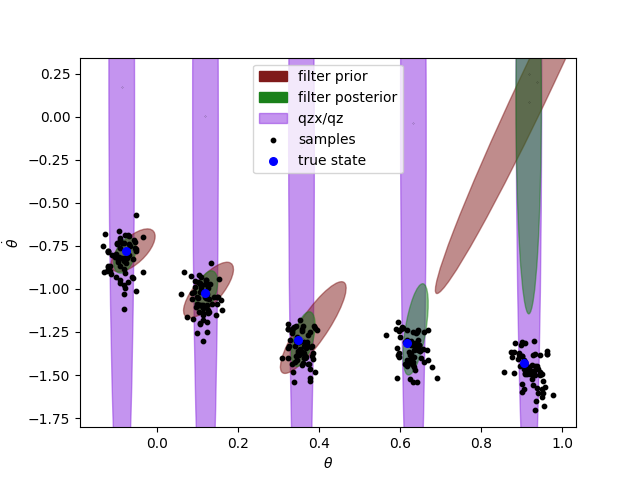}
    }
    \caption{With fixed dynamics parameters $\dyn$ and setting $p_\theta(z_1) = p_\Data(z_1)$, we demonstrate (a) the ability to learn a reasonable filter for $\theta$, $\dot{\theta}$ from image observations alone purely through matching the $p_\theta(z_1)$ prior and (b) higher accuracy filtering with additional linear measurement given by $x^{(2)}_t = \theta_t$ at training time}
    \label{fig:filter}
\end{figure}

\ifdefined\extended
\subsection{Qualitative Results}
\else
\noindent\textbf{Qualitative results:}
\fi
The trained $q_\infer(\state_t|\meas_t)$ models and corresponding filter covariances visualized in Figure \ref{fig:filter} for the pendulum environment demonstrate the ability to learn independence of the image observations and the hidden state consistent with a given dynamics model, as well as the ability to approximately recover the true underlying state simply by matching the prior $p_\theta(z_1)$. In Figure \ref{fig:reconst} we additionally show that the reconstructions of the samples from the smoothing distribution $q_\infer(z_{1:T}|x_{1:T}, u_{1:T})$ visualize the same dynamics as the sample image sequence.

\ifdefined\extended
\begin{table}[t]
\input{table.tex}
\end{table}
\fi

\ifdefined\extended
\subsection{Partial State Supervision}
\else
\noindent\textbf{Partial State Supervision:}
\fi
An advantage of our approach over prior work is the ability to use a subset of the training-time observations without compromising the correctness of the learned model. By introducing a direct linear measurement for a subset of the state components, we can perform partial state supervision. We consider the partially supervised setting with measurements $x_t^{(j)} = \theta$ and $x_t^{(j)} = \begin{bmatrix}x & y\end{bmatrix}$ for the pendulum and Airsim environments, respectively, as well as the fully supervised setting with measurements $x_t^{(j)} = \begin{bmatrix} \theta & \dot{\theta}\end{bmatrix}^\top$ and $x_t^{(j)} = \begin{bmatrix} x & y & \dot{x} & \dot{y}\end{bmatrix}^\top$ for the pendulum and Airsim environments, respectively.  In all cases, we set the measurement noise covariance $\Sigma_x = 0.05I$. As shown in Table \ref{table:mse}, using even partial supervision dramatically reduces the filtering error. This improvement is also reflected in the better latent space structures visualized in Figure \ref{fig:latent}, where the partially supervised models have a scale more similar to that of the true states (in this case a $\theta$ range of $(-\pi, \pi)$ and an $x, y$ range of $(-4, 4)$ for the pendulum and Blocks environments respectively). The effect of the improved latent state structure is also evident in the more accurate filtering trajectories shown in \ref{fig:traj}.

\ifdefined\extended
\else
\begin{wrapfigure}[14]{r}{.4\textwidth}
    \includegraphics[width=.39\textwidth]{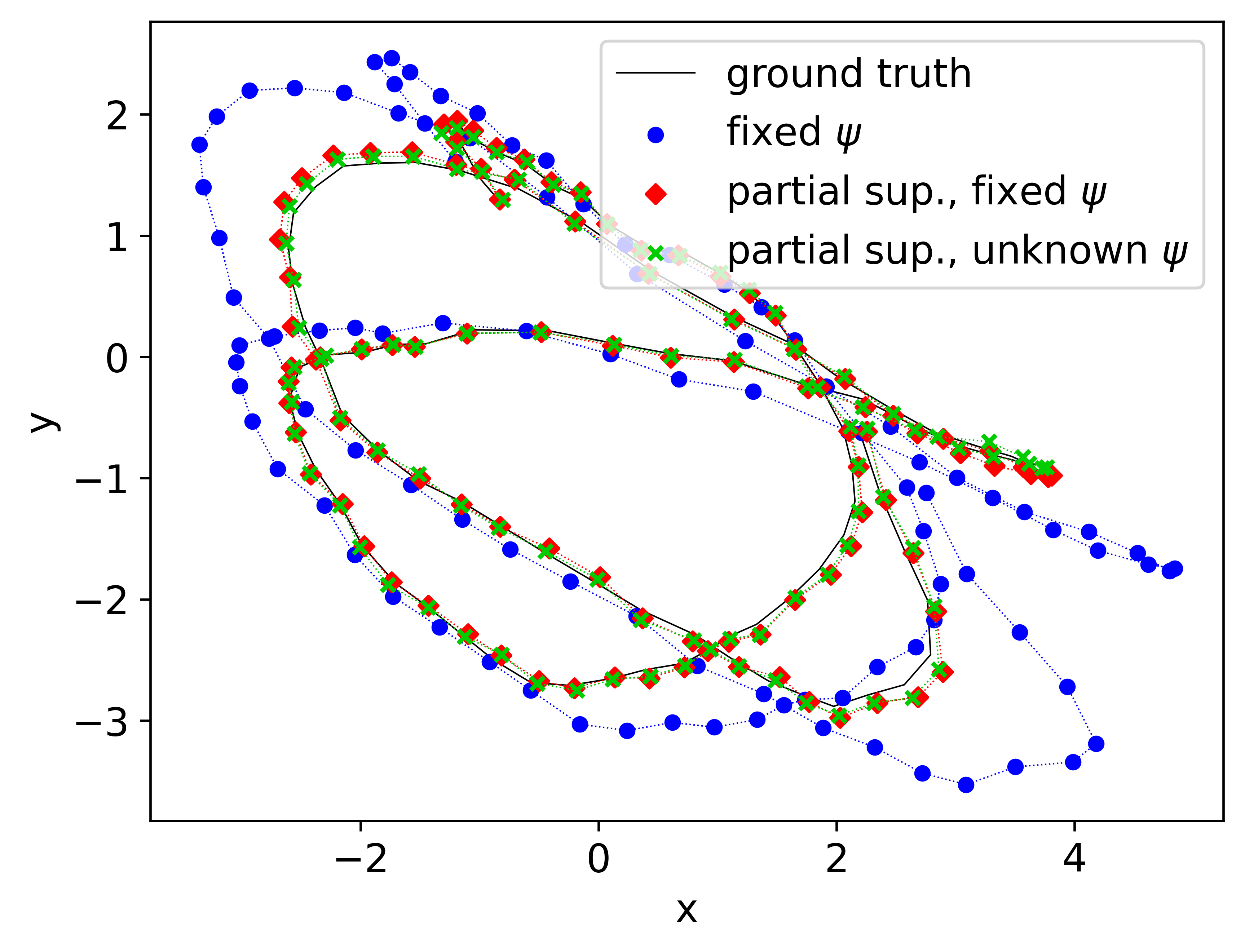}
    \caption{Filtering trajectories over $T = 100$ time steps in the Blocks environment for several different training configurations. Only pixels are used by the filter.}
    \label{fig:traj}
\end{wrapfigure}
\fi

\ifdefined\extended
\subsection{Extended Trajectory Experiments}\label{sec:extended_traj}
\else
\noindent\textbf{Extended Trajectory Experiments:}
\fi
To show that the resulting models learn a global state suitable for filtering on real-time systems, we considered the effect of evaluating the trained filtering model $q_{\infer}(\state_t|\Meas_{1:t}, \inp_{1:t})$ on trajectories of length $T' = 200$ compared to our training time trajectory length of $T = 4$ and $T = 5$ for the Airsim and Pendulum environments respectively. Table \ref{table:mse} and Figure \ref{fig:traj} demonstrate accurate filtering on the extended trajectories, even with unknown $\psi$ given at least partial supervision. The minimal degredation under unknown $\psi$ given supervision is notable as it suggests proper inference of the hidden state dynamics.

\ifdefined\extended
\begin{figure}[t!]
    \centering
    \begin{minipage}{0.48\textwidth}
    \vspace{0.7cm}
    \includegraphics[width=\textwidth]{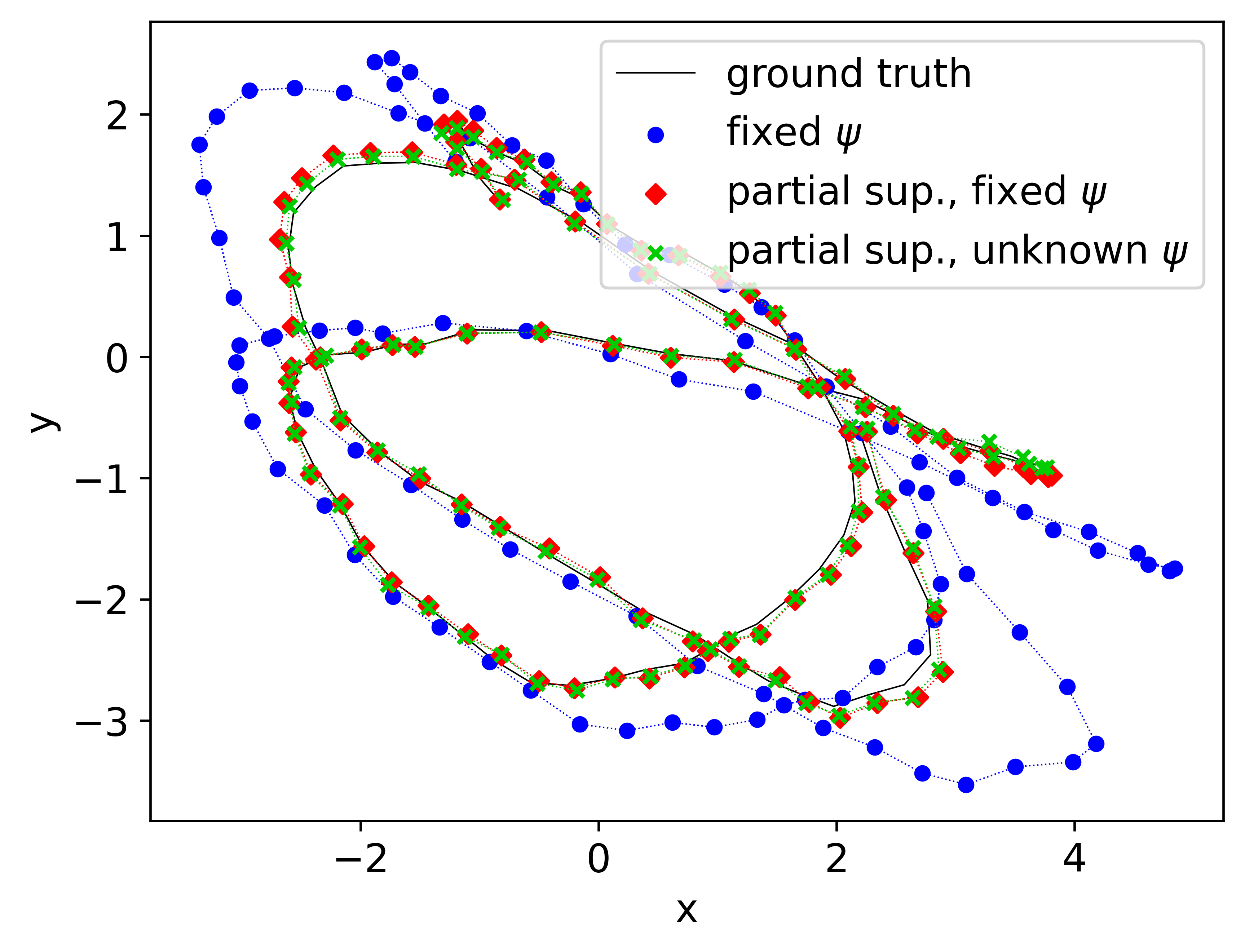}
    \caption{Filtering trajectories over $T = 100$ time steps in the Blocks environment for several different training configurations. Only pixels are used by the filter.}
    \label{fig:traj}
    \end{minipage}
    \hfill
    \begin{minipage}{0.48\textwidth}
    \centering
\subfigure[][l]{
    \includegraphics[width=95px]{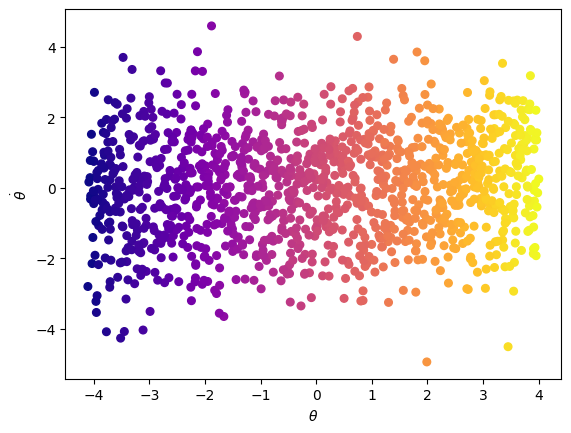}
}~\subfigure[][l]{
    \includegraphics[width=95px]{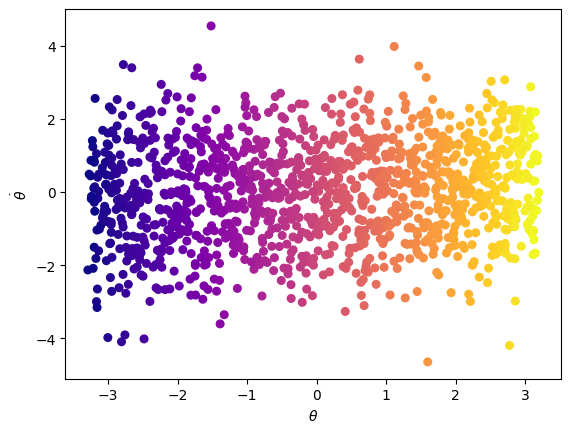}
} \\
\subfigure[][l]{
    \includegraphics[width=95px]{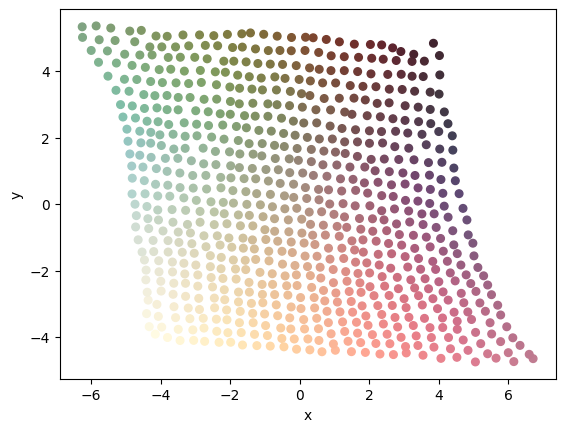}
}~\subfigure[][l]{
    \includegraphics[width=95px]{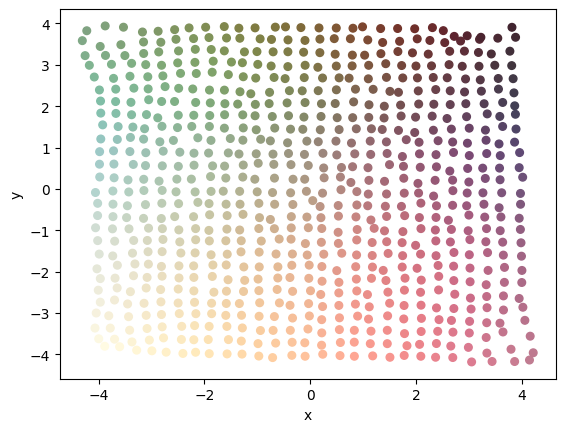}
}
    \caption{Visualizations of the latent space for the pendulum, (a) using only image data and (b) partially supervised, as well as the $x, y$ of the Block environment, (c) from images and (d) partially supervised. Note that latent state spaces for partially supervised models are more consistent with the true scale in $\theta$ and $x, y$ of $(-\pi, \pi)$ and $(-4, 4)$ respectively.}
    \label{fig:latent}
    \end{minipage}
\end{figure}
\fi
\ifdefined\extended
\subsection{Implementation Details}
\else
\noindent\textbf{Implementation Notes:}
\fi

All experiments were performed using Jax \citep{jax2018github} in conjunction with Haiku \citep{haiku2020github} and Optax \citep{optax2020github}. The code for generating the datasets and training the corresponding models is available online. \footnote{The codebase for these experiments is available at \\ \url{https://github.com/pfrommerd/variational_state_space_models}}
\ifdefined\extended

For the nonlinear models described in  \S\ref{sec:nonlinear_obs}, we used a modified version of the DCGAN \citep{radford2015unsupervised} discriminator and generator architectures for the encoder and decoder networks respectively. For the encoder, we used an output dimensionality of $32$ for the DCGAN discriminator and fed this into 3 x GELU \citep{hendrycks2020gaussian} activation + Linear layers (with 32 hidden variables) layers before reshaping the final output to be the correct shape for $r_h$.
To ensure training stability of the resulting model under Assumption \ref{assum:infer}, $ q_\infer(z_t|\meas_t^{(j)})/p_\theta(z_t)$ must be proportional to a valid probability distribution. Therefore we must enforce that $r_H(x_t)^{-1} - \Sigma_z^{-1} \succeq 0$ for all $x_t$. In our implementation for simplicity we parameterize $r_H(\meas_t)^{-1}$ by the constant function $(L^\top L + \epsilon I)^{-1} + \Sigma_z^{-1}$ for a learned matrix $L \in \R^{n\times n}$ and $\epsilon = 0.0001$.

\else
Details on the training setup and network architectures for these results can be found in the extended version \citep{extended}. 
\fi

\ifdefined\extended \else
\begin{table}
\input{table.tex}
\end{table}
\fi

\section{Conclusion}
We introduced VSSF, a new family of VAE-based models that learn low dimensional state space representations and dynamics from high dimensional observation sequences, enabling real-time filtering over the latent state. Future work will look to extend the approach to beyond globally linear latent state space dynamics.%, potentially incorporating recent advancements made in Deep-Learning aided Kalman smoothing \citep{ni2021rtsnet}.

\section*{Acknowledgements}
The authors thank Kostas Daniilidis for many helpful discussions as well as Bernadette Butcher for providing comments on the manuscript.  NM is funded by NSF awards CPS-2038873, CAREER award ECCS-2045834, and a Google Research Scholar award.  Work done by DP was partially funded by a Penn Grant for Faculty Mentoring Undergraduate Research (GFMUR).

\bibliography{references.bib}

\ifdefined\extended

\clearpage 

\appendix
\input{tex/appendix}
\fi

\end{document}

%% file: tikz/dep_rce.tikz
\begin{tikzpicture}[scale=0.85]
    \node[shape=circle,draw=black] (u1) at (1,1) {$\inp_1$};
    \node[shape=circle,draw=black] (u2) at (3,1) {$\inp_2$};

    \node[shape=circle,draw=black] (z1) at (0,0) {$\state_1$};
    \node[shape=circle,draw=black] (z2) at (2,0)
    {$\state_2$};
    \node[shape=circle,draw=black] (z3) at (4,0)
    {$\state_3$};
    
    \node[shape=circle,draw=black] (x1) at (0,-2) {$\meas_1$};
    \node[shape=circle,draw=black] (x2) at (2,-2) {$\meas_2$};
    \node[shape=circle,draw=black] (x3) at (4,-2) {$\meas_3$};
    
    \path [->] (u1) edge[bend right] node[left] {} (x2);
    \path [->] (u2) edge[bend right] node[left] {} (x3);
    
    \path [->] (z1) edge node[left] {} (x1);
    \path [->] (z2) edge node[left] {} (x2);
    \path [->] (z3) edge node[left] {} (x3);

    \path [->] (x1) edge node[left] {} (x2);
    \path [->] (x2) edge node[left] {} (x3);
    
    \path [dashed,->] (x1) edge[bend left] node[below] {} (z1);
    \path [dashed,->] (z2) edge node[right] {} (z1);
    \path [dashed,->] (x2) edge[bend right] node[left] {} (z2);
    \path [dashed,->] (u1) edge node[left] {} (z2);

\end{tikzpicture}

%% file: tikz/dep_planet.tikz
\begin{tikzpicture}[scale=0.85]
    \node[shape=circle,draw=black] (u1) at (1,1) {$\inp_1$};
    \node[shape=circle,draw=black] (u2) at (3,1) {$\inp_2$};

    \node[shape=circle,draw=black] (z1) at (0,0) {$\state_1$};
    \node[shape=circle,draw=black] (z2) at (2,0)
    {$\state_2$};
    \node[shape=circle,draw=black] (z3) at (4,0)
    {$\state_3$};
    \path [->] (z1) edge node[left] {} (z2);
    \path [->] (z2) edge node[left] {} (z3);
    
    \path [->] (u1) edge node[left] {} (z2);
    \path [->] (u2) edge node[left] {} (z3);
    
    \node[shape=circle,draw=black] (x1) at (0,-2) {$\meas_1$};
    \node[shape=circle,draw=black] (x2) at (2,-2) {$\meas_2$};
    \node[shape=circle,draw=black] (x3) at (4,-2) {$\meas_3$};
    \path [->] (z1) edge node[left] {} (x1);
    \path [->] (z2) edge node[left] {} (x2);
    \path [->] (z3) edge node[left] {} (x3);

    \path [dashed,->] (z1) edge[bend left] node[left] {} (z2);

    \path [dashed,->] (x2) edge[bend left] node[below] {} (z2);
    \path [dashed,->] (u1) edge[bend left] node[below] {} (z2);
\end{tikzpicture}

%% file: tikz/dep_our.tikz
\begin{tikzpicture}[scale=0.85]
    \node[shape=circle,draw=black] (u1) at (1,1) {$\inp_1$};
    \node[shape=circle,draw=black] (u2) at (3,1) {$\inp_2$};

    \node[shape=circle,draw=black] (z1) at (0,0) {$\state_1$};
    \node[shape=circle,draw=black] (z2) at (2,0)
    {$\state_2$};
    \node[shape=circle,draw=black] (z3) at (4,0)
    {$\state_3$};
    \path [->] (z1) edge node[left] {} (z2);
    \path [->] (z2) edge node[left] {} (z3);
    
    \path [->] (u1) edge node[left] {} (z2);
    \path [->] (u2) edge node[left] {} (z3);
    
    \node[shape=circle,draw=black] (x1) at (0,-2) {$\meas_1$};
    \node[shape=circle,draw=black] (x2) at (2,-2) {$\meas_2$};
    \node[shape=circle,draw=black] (x3) at (4,-2) {$\meas_3$};
    \path [->] (z1) edge node[left] {} (x1);
    \path [->] (z2) edge node[left] {} (x2);
    \path [->] (z3) edge node[left] {} (x3);

    \path [dashed,->] (z1) edge[bend left] node[left] {} (z2);
    \path [dashed,->] (x2) edge[bend left] node[below] {} (z2);
    \path [dashed,->] (x3) edge node[below] {} (z2);
    \path [dashed,->] (u1) edge[bend left] node[above] {} (z2);
    \path [dashed,->] (u2) edge node[above] {} (z2);
\end{tikzpicture}

%% file: tikz/flow.tikz
\usetikzlibrary{shapes.geometric}
\definecolor{enc_color}{rgb}{0.81,0.89,0.95}
\definecolor{prop_color}{rgb}{0.9,0.9,0.9}
\definecolor{update_color}{rgb}{0.9,0.72,0.69}
\definecolor{dec_color}{rgb}{1, 0.9, 0.6}

\def\mspace{6}
\def\yrowf{-2.4}
\def\yrows{-3.9}
\def\yrowt{-5.2}

\begin{tikzpicture}[
net/.style={
  trapezium, trapezium angle=75, draw,inner xsep=0.2cm,
  inner ysep=0.5cm, outer sep=0pt, rounded corners=0.1cm,
  minimum height=1.81mm
},prop/.style={
  rectangle, draw,inner xsep=0.1cm,
  inner ysep=0.1cm, outer sep=0pt, rounded corners=0.1cm,
  minimum height=1.81mm,
  fill=prop_color
},update/.style={
  rectangle, draw,inner xsep=0.1cm,
  inner ysep=0.1cm, outer sep=0pt, rounded corners=0.1cm,
  minimum height=1.81mm,
  fill=update_color
},bprop/.style={
  rectangle, draw,inner xsep=0.1cm,
  inner ysep=0.1cm, outer sep=0pt, rounded corners=0.1cm,
  minimum height=1.81mm,
  fill=prop_color
}]
    \node (rm1) at (0, 1.3) {$\Meas_1$};
    \node (rm2) at (\mspace, 1.3) {$\Meas_2$};
    \node (rm3) at (2*\mspace, 1.3) {$\Meas_3$};

    \node[net, fill=enc_color, shape border rotate=180] (m1) at (0,0) {Enc};
    \node[net, fill=enc_color, shape border rotate=180] (m2) at (\mspace,0) {Enc};
    \node[net, fill=enc_color, shape border rotate=180] (m3) at (2*\mspace,0) {Enc};
    
    \draw [-latex] (rm1) -- (m1);
    \draw [-latex] (rm2) -- (m2);
    \draw [-latex] (rm3) -- (m3);
    
    \node (e1) at (0, -1.2) {$q(\state_1|\Meas_1)$};
    \node (e2) at (\mspace, -1.2) {$q(\state_2|\Meas_2)$};
    \node (e3) at (2*\mspace, -1.2) {$q(\state_3|\Meas_3)$};
    
    \node[prop, shape border rotate=270] (p2) at (\mspace/2, \yrowf) {$q(\state_2|\Meas_1,\inp_1)$};

    \coordinate (e1j) at (0, \yrowf);
    \draw [-latex] (e1) -- (e1j) |- (p2.west);

    \node[update, shape border rotate=270] (u2) at (\mspace, \yrowf) {$q(\state_2|\Meas_{1:2},\inp_1)$};
    
    \draw [-latex] (p2) -- (u2);
    \draw [-latex] (e2) -- (u2);

    \node[prop, shape border rotate=270] (p3) at (\mspace + \mspace/2, \yrowf) {$q(\state_3|\Meas_{1:2},\inp_{1:2})$};
    
    \draw [-latex] (u2) -- (p3);
    
    \node[update, shape border rotate=270] (u3) at (2*\mspace, \yrowf) {$q(\state_3|\Meas_{1:3},\inp_{1:2})$};
    
    \draw [-latex] (p3) -- (u3);
    \draw [-latex] (e3) -- (u3);

    \node[rotate=90] at (2*\mspace, \yrowf/2 + \yrows/2) {$\sim$};
    \node (z3) at (2*\mspace, \yrows) {$\state_3$};

    \node[bprop, shape border rotate=270] (bp3) at (\mspace + \mspace/2, \yrows) {$q(\state_2|\state_3,\Meas_{1:2},\inp_{1:2})$};
    
    \draw [-latex] (z3) -- (bp3.east);
    
    \coordinate (u2j1) at (\mspace, \yrowf/2 + \yrows/2);
    \coordinate (u2j2) at (\mspace + \mspace/2, \yrowf/2 + \yrows/2);
    
    \draw [-latex] (u2.south) -- (u2j1) -- (u2j2) -- (bp3.north);

    \node at (\mspace + 0.7, \yrows) {$\sim$};
    \node (z2) at (\mspace, \yrows) {$\state_2$};

    \node[bprop, shape border rotate=270] (bp2) at (\mspace/2, \yrows) {$q(\state_1|\state_2,\Meas_{1},\inp_1)$};
    
    \draw [-latex] (z2) -- (bp2);
    
    \coordinate (e1j2) at (0, \yrowf/2 + \yrows/2);
    \coordinate (e1j3) at (\mspace/2, \yrowf/2 + \yrows/2);
    \draw [-latex] (e1j) -- (e1j2) -- (e1j3) -- (bp2.north);

    \node at (0.7, \yrows) {$\sim$};
    \node (z1) at (0, \yrows) {$\state_1$};

    % The decoder
    \node[net, fill=dec_color] (d1) at (0,\yrowt) {Dec};
    \node[net, fill=dec_color] (d2) at (\mspace,\yrowt) {Dec};
    \node[net, fill=dec_color] (d3) at (2*\mspace,\yrowt) {Dec};
    
    \draw [-latex] (z1) -- (d1);
    \draw [-latex] (z2) -- (d2);
    \draw [-latex] (z3) -- (d3);

    \node at (0, \yrowt -1.2) {$p(\Meas_1|\state_1)$};
    \node at (\mspace, \yrowt -1.2) {$p(\Meas_2|\state_2)$};
    \node at (2*\mspace, \yrowt -1.2) {$p(\Meas_3|\state_3)$};

    % Put "forward pass" and "backward pass" labels
    \node at (2*\mspace + 3, \yrowf) {Forward Pass};
    \node at (2*\mspace + 3, \yrows) {Backward Pass};
\end{tikzpicture}

%% file: table.tex
\centering
    \begin{tabular}{@{}lllllllll@{}}
\toprule
Supervision & \multicolumn{2}{c}{None} & \phantom{abs} & \multicolumn{2}{c}{Partial} & \phantom{abs} & \multicolumn{2}{c}{Full} \\
 \cmidrule{2-3} \cmidrule{5-6} \cmidrule{8-9}
 & $\dyn$ fixed & $\dyn$ unknown & & $\dyn$ fixed & $\dyn$ unknown & & $\dyn$ fixed & $\dyn$ unknown \\
Pendulum & 0.0458 & 0.558* && 0.000452 & 0.000369 && 0.000211 & 0.000368 \\
Blocks & 0.244 & 9.715* && 0.00292 & 0.00488 & & 0.00193 & 0.00230 \\
Zhangjiajie & 2.308 & 6.186* && 0.0519 & 0.137 && 0.0089 & 0.0413 \\
\bottomrule
\end{tabular}
\caption{The mean squared position/angle error between the filtering posterior mean and the ground truth for different models on extended trajectories of length 200. For models where $\psi$ is optimized, we use the learned $\psi$ to evaluate the filter. (*) Note that these models are all degenerate.}
\label{table:mse}

%% file: tex/appendix.tex
\section{Supplementary Proofs}
\label{appendix:model}

\subsection{Evidence Lower Bound}
We use the same general ELBO structure as \cite{kingma2014autoencoding}, adapted for our problem domain. For completeness we reproduce the derivation of the ELBO below.
\begin{proposition}
The Evidence Lower Bound $\L(\gener, \infer, \dyn, \Meas_{1:T}, \inp_{1:T-1}) \leq \log p_{\gener, \dyn}(\Meas_{1:T}| u_{1:T-1})$ given by
$$\L(\gener, \infer, \dyn, \Meas_{1:T}, \inp_{1:T-1}) = \log p_{\gener, \dyn}(\Meas_{1:T}| u_{1:T-1}) - G,$$
where $G = \DKL(q_{\infer, \dyn}(\state_{1:T}|\Meas_{1:T}, \inp_{1:T-1}) || p_{\gener, \dyn}(\state_{1:T}|\Meas_{1:T}, \inp_{1:T-1}))$ can be written
\begin{align*}
\L(\gener, \infer, \dyn, \Meas_{1:T}, \inp_{1:T-1}) &= -\DKL(q_{\infer, \dyn}(\state_{1:T}|\Meas_{1:T}, \inp_{1:T-1}) || p_{\theta, \psi}(\state_{1:T}|\inp_{1:T-1}))) \\
    &\hspace{0.3cm} + \E_{q_{\infer, \dyn}(\state_{1:T}|\Meas_{1:T}, \inp_{1:T-1})}[\log p_{\gener, \dyn}(\Meas_{1:T}|\state_{1:T})].
\end{align*}
\\
\noindent Proof. Using $\E_{q_{\infer, \dyn}}$ as shorthand for $\E_{q_{\infer, \dyn}(\state_{1:T}|\Meas_{1:T}, \inp_{1:T-1})}$.
\begin{align*}
\L(\gener, \infer, \dyn, \Meas_{1:T}, \inp_{1:T-1}) &= - \DKL(q_{\infer, \dyn}(\state_{1:T}|\Meas_{1:T}, \inp_{1:T-1}) || p_{\gener, \dyn}(\state_{1:T}|\Meas_{1:T}, \inp_{1:T-1})) \\
&\hspace{1.5cm} + \log p_{\gener, \dyn}(\Meas_{1:T}| u_{1:T-1})  \\
&= \E_{q_{\infer, \dyn}}[-\log q_{\infer, \dyn}(\state_{1:T}|\Meas_{1:T}, \inp_{1:T-1}) + \log p_{\gener, \dyn}(\Meas_{1:T}| u_{1:T-1}) \\
&\hspace{1.5cm} + \log p_{\gener, \dyn}(\state_{1:T}|\Meas_{1:T}, \inp_{1:T-1})] \\
&= \E_{q_{\infer, \dyn}}[-\log q_{\infer, \dyn}(\state_{1:T}|\Meas_{1:T}, \inp_{1:T-1}) + \log p_{\gener, \dyn}(\state_{1:T},\Meas_{1:T}| \inp_{1:T-1})] \\
&= \E_{q_{\infer, \dyn}}[-\log q_{\infer, \dyn}(\state_{1:T}|\Meas_{1:T}, \inp_{1:T-1}) + \log p_{\gener, \dyn}(\state_{1:T}|\inp_{1:T-1}) \\
&\hspace{1.5cm} + \log p_{\gener, \dyn}(\Meas_{1:T}|\state_{1:T-1})] \\
&= -\DKL(q_{\infer, \dyn}(\state_{1:T}|\Meas_{1:T}, \inp_{1:T-1})||p_{\gener, \dyn}(\state_{1:T}|\inp_{1:T-1})] + \E_{q_{\infer, \dyn}}[\log p_{\gener, \dyn}(\Meas_{1:T}|\state_{1:T-1})]
\end{align*}
\end{proposition}

\subsection{Factorization of $q(\state_t|\state_{t + 1}, \Meas_{1:t}, \inp_{1:t})$}

The factorization of $q(\state_t|\state_{t + 1}, \Meas_{1:t}, \inp_{1:t})$ follows from a straightforward application of Bayes' rule.

\begin{align*}
    q(\state_t|\state_{t + 1}, X_{1:t}, \inp_{1:t}) &= \frac{q(\state_{t + 1}|\state_t, \Meas_{1:T}, \inp_{1:t}) q(\state_t|\Meas_{1:t}, \inp_{1:t})}{q(\state_{t + 1}|\Meas_{1:t},\inp_{1:t})}
\intertext{Since $\int q(\state_t|\state_{t + 1}, X_{1:t}, \inp_{1:t}) d\state_t = 1$, the denominator $q(\state_{t + 1}|\Meas_{1:t},\inp_{1:t})$ is just a normalization constant. Therefore}
    q(\state_t|\state_{t + 1}, X_{1:t}, \inp_{1:t}) &\propto q(\state_{t + 1}|\state_t, \Meas_{1:T}, \inp_{1:t}) q(\state_t|\Meas_{1:t}, \inp_{1:t})
\end{align*}

\subsection{Proof of Correctness for Using the Modified Prior}\label{append:correctness}
\begin{theorem}\label{thm:correctness} Provided Assumption \ref{assum:infer} holds, maximizing the ELBO (\ref{eqn:elbo}) over $p_\Data(\Meas_{1:T}| \inp_{1:T-1})$ under the modification that $q_{\theta, \dyn}(z_t) := p_{\theta, \dyn}(z_t)$ in the posterior update equation (\ref{eqn:update}),
\begin{align*}
    q_{\infer,\dyn}(z_t|\Meas_{1:t}, u_{1:t-1}) \propto q_{\infer, \dyn}(z_t|\Meas_{1:t-1},u_{1:t-1})\prod_{j=1}^k \frac{q_\infer(z_t|\Meas_t^{(j)})}{q_{\infer,\dyn}(z_t)},
\end{align*}
is equivalent to maximizing the original ELBO over $p_\Data(\Meas_{1:T}| \inp_{1:T-1})$.
\\

\noindent Proof. Let $\modified{q}_{\infer, \dyn}(z_{1:T}|\Meas_{1:T}, u_{1:T-1})$ denote the smoothing distribution $q_{\infer, \dyn}(z_{1:T}|\Meas_{1:T}, u_{1:T-1})$ as described in \S\ref{sec:formulation} with $q_{\infer, \dyn}(z_t)$ replaced by $p_{\theta, \dyn}(z_t)$ in Equation (\ref{eqn:update}) and $\modified{\L}(\infer, \gener, \dyn, \Meas_{1:T}, \inp_{1:T-1})$ denote the resulting modified ELBO. 

First we show any optimal $\hat{\gener}, \hat{\dyn}, \hat{\infer}$ that maximizes the true ELBO $\E_{\Meas_{1:T}, \inp_{1:T-1}}\L(\cdot, \Meas_{1:T}, \inp_{1:T-1})$ is an optimal solution for the modified ELBO $\E_{\Meas_{1:T}, \inp_{1:T-1}}\modified{\L}(\cdot, \Meas_{1:T}, \inp_{1:T-1})$. For notational brevity we will omit expectations over $\Meas_{1:T}, \inp_{1:T-1}$.

Note that under Assumption \ref{assum:infer}, for parameters $\hat{\infer},\hat{\dyn}$ maximizing the true ELBO, the latent priors of the inference and generative distributions are equivalent, i.e $q_{\hat{\infer}, \hat{\dyn}}(z_t) = p_{\theta, \dyn}(z_t)$. Therefore for any optimal $\hat{\infer}, \hat{\dyn}$ for the original ELBO, the modified and unmodified inference distributions are the same, i.e  $\modified{q}_{\hat{\infer}, \hat{\dyn}}(z_{1:T}|\Meas_{1:T}, u_{1:T-1}) = q_{\hat{\infer}, \hat{\dyn}}(z_{1:T}|\Meas_{1:T}, u_{1:T-1})$.
Since the modified ELBO is given by
\begin{align*}
    \modified{\L}(\infer, \gener, \dyn, \Meas_{1:T},\inp_{1:T-1}) &= \log p_{\gener, \dyn}(\Meas_{1:T}| \inp_{1:T-1}) \\
    &\hspace{5mm} -\DKL(\modified{q}_{\infer, \dyn}(\state_{1:T}|\Meas_{1:T},\inp_{1:T-1})\, ||\, p_{\gener, \dyn}(\state_{1:T}|\Meas_{1:T},\inp_{1:T-1})),
\end{align*}
under optimal $\hat{\infer}, \hat{\dyn}, \hat{\gener}$ for $\L$, the modified ELBO becomes
\begin{align*}
    \modified{\L}(\hat{\infer}, \hat{\gener}, \hat{\dyn}, \Meas_{1:T},\inp_{1:T-1}) &= \log p_{\hat{\gener}, \hat{\dyn}}(\Meas_{1:T}| \inp_{1:T-1}) \\
    &\hspace{5mm} -\DKL(q_{\hat{\infer}, \hat{\dyn}}(\state_{1:T}|\Meas_{1:T},\inp_{1:T-1})\, ||\, p_{\hat{\gener}, \hat{\dyn}}(\state_{1:T}|\Meas_{1:T},\inp_{1:T-1})).
\end{align*}
By Assumption \ref{assum:infer} the KL divergence term is zero and $\hat{\gener}, \hat{\dyn}$ maximize $\log p_{\hat{\gener}, \hat{\dyn}}(\Meas_{1:T}, \inp_{1:T-1})$, so $\hat{\infer}, \hat{\dyn}, \hat{\gener}$ must also maximize $\modified{\L}$.

Now consider any $\modified{\gener}, \modified{\infer}, \modified{\dyn}$ that maximize $\modified{\L}$. By the existence of $\hat{\gener}, \hat{\infer}, \hat{\dyn}$, in order for $\modified{\gener}, \modified{\infer}, \modified{\dyn}$ to maximize $\modified{\L}$, it must be that $$\modified{\L}(\modified{\infer}, \modified{\gener}, \modified{\dyn}, \Meas_{1:T}, u_{1:T-1}) \geq \modified{\L}(\hat{\infer}, \hat{\gener}, \hat{\dyn}, \Meas_{1:T}, u_{1:T-1}) = \log p_{\hat{\gener}, \hat{\dyn}}(\Meas_{1:T}| \inp_{1:T-1}).$$ 
Note that $\hat{\gener}, \hat{\dyn}$ maximize $p_{\gener, \dyn}(\Meas_{1:T}| \inp_{1:T-1})$. Since the KL divergence is non-negative this implies that
$$\log p_{\modified{\gener}, \modified{\dyn}}(\Meas_{1:T}| \inp_{1:T-1}) = \log p_{\hat{\gener}, \hat{\dyn}}(\Meas_{1:T}| \inp_{1:T-1}),$$
and
$$\esssup_{\Meas_{1:T}, u_{1:T-1} \sim p_{\Datadist}(\Data)}\DKL(\modified{q}_{\modified{\infer}, \modified{\dyn}}(\state_{1:T}|\Meas_{1:T},\inp_{1:T-1})\, ||\, p_{\modified{\gener}, \modified{\dyn}}(\state_{1:T}|\Meas_{1:T},\inp_{1:T-1})) = 0$$
Therefore the inferred latent state prior matches the generative latent state prior, i.e $\modified{q}_{\modified{\infer}, \modified{\dyn}}(z_t) = p_{\modified{\gener}, \modified{\dyn}}(z_t)$, and consequently $\modified{q}_{\modified{\infer}, \modified{\dyn}}(z_{1:T}|\Meas_{1:T}, u_{1:T-1}) = q_{\modified{\infer}, \modified{\dyn}}(z_{1:T}|\Meas_{1:T}, u_{1:T-1})$, implying that $\modified{\gener}, \modified{\infer}, \modified{\dyn}$ also maximize the true ELBO.
\end{theorem}

\subsection{Correctness of learned filter $q(\state_{t}|\Meas_{1:t},\inp_{1:t-1})$ and inference model $q(\state_t|\meas^{(j)}_{t})$}
\label{append:correct_filter}
Assumption \ref{assum:infer} guarantees that parameters $\hat{\gener},\hat{\infer},\hat{\dyn}$ which maximize $\E_{\Meas_{1:T},u_{1:T-1}}\L(\cdot, \Meas_{1:T}, u_{1:T-1})$ parameterize the proper smoothing inference distribution, i.e for all $\Meas_{1:T}, u_{1:T-1} \sim p_{\Datadist}(\Meas_{1:T}, u_{1:T-1})$,
$$q_{\hat{\infer}, \hat{\dyn}}(z_{1:T}|\Meas_{1:T}, u_{1:T-1}) = p_{\hat{\gener},\hat{\dyn}}(z_{1:T}|\Meas_{1:T}, u_{1:T-1}).$$
This is not necessarily sufficient to guarantee that either
$q_{\hat{\infer}, \hat{\dyn}}(z_{t}|\Meas_{1:t}, u_{1:t-1}) = p_{\hat{\gener},\hat{\dyn}}(z_{t}|\Meas_{1:t}, u_{1:t-1})$ or $q_{\hat{\infer},\hat{\dyn}}(z_t|\meas_{t}^{(j)})=p_{\hat{\gener}}(z_t|\meas_t^{(j)})$ everywhere.

In order for the inferred filtering distribution to satisfy $q_{\hat{\infer}, \hat{\dyn}}(z_{t}|\Meas_{1:t}, u_{1:t-1}) = p_{\hat{\gener},\hat{\dyn}}(z_{t}|\Meas_{1:t}, u_{1:t-1})$, we must additionally have that the distribution $q_{\dyn}(z_{t+1}|z_t,u_t)$ used in the time-reversed decomposition \eqref{eqn:factor} satisfies (i) the dynamics model used by the inference distribution $q_{\infer,\dyn}(z_{1:T}|\Meas_{1:T},\inp_{1:T-1})$ and the prior $p_{\gener,\dyn}(z_{1:T},\inp_{1:T-1})$ must be the same such that $q_\dyn(z_{t+1}|z_t,u_t) = p_\dyn(z_{t+1}|z_t,u_t)$ and (ii) $q_{\dyn}(z_{t+1}|z_t,u_t)$ must be nonzero for all $z_t,z_{t+1},u_t$. We satisfy (i) by explicitly setting $q_\dyn(z_{t+1}|z_t,u_t) = p_\dyn(z_{t+1}|z_t,u_t)$ in our decomposition given in \S\ref{sec:formulation} and note that (ii) is satisfied by the multivariate Gaussian parameterization for L-VSSF given in \S\ref{sec:lvssm}. Under both of these conditions

\begin{align*}
    q_{\hat{\infer}, \hat{\dyn}}(z_{1:T}|\Meas_{1:T}, u_{1:T-1}) &= p_{\hat{\gener},\hat{\dyn}}(z_{1:T}|\Meas_{1:T}, u_{1:T-1}).
\intertext{Using equality, under the reverse decomposition given by Equation \eqref{eqn:bkw}}
    \Longrightarrow\quad q_{\hat{\infer}, \hat{\dyn}}(z_{t}|z_{t+1},\Meas_{1:t}, u_{1:t}) &= p_{\hat{\gener}, \hat{\dyn}}(z_{t}|z_{t+1},\Meas_{1:t}, u_{1:t}) \quad \forall \,  t \in [0, T]\\
    \Longrightarrow\quad q_{\hat{\dyn}}(z_{t+1}|z_{t},u_{t})q_{\hat{\infer},\hat{\dyn}}(z_{t}|\Meas_{1:t},\inp_{1:t-1}) &\propto p_{\hat{\dyn}}(z_{t+1}|z_{t},u_{t})p_{\hat{\gener},\hat{\dyn}}(z_{t}|\Meas_{1:t},\inp_{1:t-1}) \quad \forall \,  t \in [0, T].
\intertext{Using conditions (i) and (ii) we can conclude that the filtering distributions must be identical for all $t$}
    \Longrightarrow\quad q_{\hat{\infer},\hat{\dyn}}(z_{t}|\Meas_{1:t},\inp_{1:t-1}) &= p_{\hat{\gener},\hat{\dyn}}(z_{t}|\Meas_{1:t},\inp_{1:t-1})\quad \forall \,  t \in [0, T]. \numberthis \label{eqn:correct_filter}
\end{align*}
It remains to be shown that $q_{\hat{\infer}}(\state_t| \meas_t^{(j)})$ satisfy $q_{\hat{\infer}}(\state_t| \meas_t^{(j)}) = p_{\hat{\gener}}(\state_t | \meas_t^{(j)})$. From Equations (\ref{eqn:update}), (\ref{eqn:correct_filter}) and Assumption \ref{assum:infer} it follows that
\begin{align}
    \prod_{j=1}^k q_{\hat{\infer}}(\state_t|\meas_t^{(j)}) = \prod_{j=1}^k p_{\hat{\gener}}(\state_t|\meas_t^{(j)})
\end{align}
Provided that (iii) $q_{\hat{\infer}}(\state_t| \meas_t^{(j)})$ is always nonzero and (iv) for all $k' > 1$
\begin{align*}
    \prod_{j=1,j\neq k'}^k q_{\hat{\infer}}(\state_t|\meas_t^{(j)}) = \prod_{j=1,j\neq k'}^k p_{\hat{\gener}}(\state_t|\meas_t^{(j)})
\end{align*}
it follows directly that $q_{\hat{\infer}}(\state_t| \meas_t^{(j)}) = p_{\hat{\gener}}(\state_t | \meas_t^{(j)})$.

For L-VSSF, (iii) is satisfied by the choice of multivariate Gaussian distributions and (iv) is satisfied if $k - 1$ of the observation models satisfy $q_{\infer}(\state_t|\meas_t^{(j)}) = p_{\gener}(\state_t|\meas_t^{(j)})$. This holds for the linear observation model given in \S\ref{sec:linear_obs} and (iv) holds for all experiments run in \S\ref{sec:experiment}.

We note that it is also possible to satisfy (iv) by maximizing an objective function given by the ELBO for the full distribution $\log p(\Meas_{1:T} | \inp_{1:t-1})$, in addition to an ELBO for $$\sum_{k'=2}^k \log p(\{\{x_t^{(j)}\}_{j=1,j\neq k'}^k\}_{t=1}^T | \inp_{1:t-1}),$$ i.e by inferring both the full filtering distribution, as well as each filter where the observation $j=k'$ has been removed. We leave experimental verification of this approach as a matter for future work.

\subsection{Derivation of $q(\state_t|\state_{t + 1}, X_{1:t}, \inp_{1:t})$ for L-VSSF}
\label{append:linear_model_derivation}

\begin{theorem} For the linear dynamics formulation in Section \ref{sec:lvssm} we have $q_{\varphi, \psi}(z_t|z_{t+1}, X_{1:t}, u_{1:t-1}) = \N(\ell_t, L_t)$ where
\begin{align*}
    L_t^{-1} &= P_{t|t}^{-1} + A^\top \Sigma_w^{-1}A \\
    L_t^{-1}\ell_t &= A^\top \Sigma_w^{-1}(P_{t|t}^{-1} - Bu_t) + P_{t|t}^{-1}p_{t|t}
\end{align*}

\noindent Proof.
From Equation (\ref{eqn:bkw}) it follows
\begin{align*}
    q_{\infer, \dyn}(\state_t | \state_{t + 1}, \Meas_{1:t}, \inp_{1:t}) &\propto q_{\dyn}(\state_{t + 1} | \state_t, \inp_t)q_{\infer, \dyn}(\state_t|\Meas_{1:t}, \inp_{1:t-1}) \\
    &\propto \N(\state_{t + 1}, A\state_t + B\inp_t, \Sigma_w)\N(\state_t, p_{t|t}, P_{t|t}) \\
\intertext{Where $\N(x, \mu, \Sigma)$ denotes the multivariate normal density function with mean $\mu$ and covariance matrix $\Sigma$ evaluated at $x$. Since}
 \N(\state_{t + 1}, A\state_t + B\inp_t, \Sigma_w) &\propto e^{(z_{t + 1} - (Az_{t} + Bu_t))^\top \Sigma_w^{-1} (z_{t + 1} - (Az_{t} + Bu_t))} \\
 &\propto e^{(A^{-1}z_{t + 1} - z_{t} + A^{-1}Bu_t))^\top A^\top \Sigma_w^{-1} A (A^{-1}z_{t + 1} - z_{t} + A^{-1}Bu_t)} \\
\intertext{Therefore $\N(\state_{t + 1}, A\state_t + B\inp_t, \Sigma_w) \propto \N(\state_t, A^{-1}z_{t + 1} - A^{-1}Bu_t, (A^{-1})^\top \Sigma_wA^{-1})$. Consequently}
 q_{\infer, \dyn}(\state_t | \state_{t + 1}, \Meas_{1:t}, \inp_{1:t}) &\propto \N(\state_t, A^{-1}z_{t + 1} - A^{-1}Bu_t, A^\top \Sigma_wA)\N(\state_t, p_{t|t}, P_{t|t})
\intertext{For the product of two multivariate normal density functions $g(x) = \N(x, a, A)\N(x, b, B)$ we can write $g(x) \propto \N(x, c, C)$ where $C^{-1} = A^{-1} + B^{-1}$, $C^{-1}c = A^{-1}a + B^{-1}b$ \citep{bromiley2003products}. It follows directly that}
q_{\infer, \dyn}(\state_t | \state_{t + 1}, \Meas_{1:t}, \inp_{1:t}) &\propto \N(\state_t, \ell_t, L_t)
\end{align*}
Where $\ell_t, L_t$ are defined as above. Because $q_{\infer, \dyn}(\state_t | \state_{t + 1}, \Meas_{1:t}, \inp_{1:t})$ is a distribution over $\state_t$, this is in fact an equality, completing the proof.
\end{theorem}

%% file: paper.bbl
\begin{thebibliography}{34}
\providecommand{\natexlab}[1]{#1}
\providecommand{\url}[1]{\texttt{#1}}
\expandafter\ifx\csname urlstyle\endcsname\relax
  \providecommand{\doi}[1]{doi: #1}\else
  \providecommand{\doi}{doi: \begingroup \urlstyle{rm}\Url}\fi

\bibitem[Babaeizadeh et~al.(2017)Babaeizadeh, Finn, Erhan, Campbell, and
  Levine]{babaeizadeh2017stochastic}
Mohammad Babaeizadeh, Chelsea Finn, Dumitru Erhan, Roy~H Campbell, and Sergey
  Levine.
\newblock Stochastic variational video prediction.
\newblock \emph{arXiv preprint arXiv:1710.11252}, 2017.

\bibitem[Banijamali et~al.(2018)Banijamali, Shu, Bui, Ghodsi,
  et~al.]{banijamali2018robust}
Ershad Banijamali, Rui Shu, Hung Bui, Ali Ghodsi, et~al.
\newblock Robust locally-linear controllable embedding.
\newblock In \emph{International Conference on Artificial Intelligence and
  Statistics}, pages 1751--1759. PMLR, 2018.

\bibitem[Bayer et~al.(2021)Bayer, Soelch, Mirchev, Kayalibay, and van~der
  Smagt]{bayer2021mind}
Justin Bayer, Maximilian Soelch, Atanas Mirchev, Baris Kayalibay, and Patrick
  van~der Smagt.
\newblock Mind the gap when conditioning amortised inference in sequential
  latent-variable models, 2021.

\bibitem[Bradbury et~al.(2018)Bradbury, Frostig, Hawkins, Johnson, Leary,
  Maclaurin, Necula, Paszke, Vander{P}las, Wanderman-{M}ilne, and
  Zhang]{jax2018github}
James Bradbury, Roy Frostig, Peter Hawkins, Matthew~James Johnson, Chris Leary,
  Dougal Maclaurin, George Necula, Adam Paszke, Jake Vander{P}las, Skye
  Wanderman-{M}ilne, and Qiao Zhang.
\newblock {JAX}: composable transformations of {P}ython+{N}um{P}y programs.
\newblock 2018.
\newblock URL \url{http://github.com/google/jax}.

\bibitem[Briers et~al.(2010)Briers, Doucet, and Maskell]{briers2010smoothing}
Mark Briers, Arnaud Doucet, and Simon Maskell.
\newblock Smoothing algorithms for state--space models.
\newblock \emph{Annals of the Institute of Statistical Mathematics},
  62\penalty0 (1):\penalty0 61, 2010.

\bibitem[Bromiley(2003)]{bromiley2003products}
Paul Bromiley.
\newblock Products and convolutions of gaussian probability density functions.
\newblock \emph{Tina-Vision Memo}, 3\penalty0 (4):\penalty0 1, 2003.

\bibitem[Campbell et~al.(2021)Campbell, Shi, Rainforth, and
  Doucet]{campbell2021online}
Andrew Campbell, Yuyang Shi, Thomas Rainforth, and Arnaud Doucet.
\newblock Online variational filtering and parameter learning.
\newblock \emph{Advances in Neural Information Processing Systems}, 34, 2021.

\bibitem[Child(2020)]{child2020very}
Rewon Child.
\newblock Very deep vaes generalize autoregressive models and can outperform
  them on images.
\newblock \emph{arXiv preprint arXiv:2011.10650}, 2020.

\bibitem[Denton and Fergus(2018)]{denton2018stochastic}
Emily Denton and Rob Fergus.
\newblock Stochastic video generation with a learned prior, 2018.

\bibitem[Devlin et~al.(2018)Devlin, Chang, Lee, and Toutanova]{devlin2018bert}
Jacob Devlin, Ming-Wei Chang, Kenton Lee, and Kristina Toutanova.
\newblock Bert: Pre-training of deep bidirectional transformers for language
  understanding.
\newblock \emph{arXiv preprint arXiv:1810.04805}, 2018.

\bibitem[Goodfellow et~al.(2014)Goodfellow, Pouget-Abadie, Mirza, Xu,
  Warde-Farley, Ozair, Courville, and Bengio]{goodfellow2014generative}
Ian~J. Goodfellow, Jean Pouget-Abadie, Mehdi Mirza, Bing Xu, David
  Warde-Farley, Sherjil Ozair, Aaron Courville, and Yoshua Bengio.
\newblock Generative adversarial networks, 2014.

\bibitem[Hafner et~al.(2019)Hafner, Lillicrap, Fischer, Villegas, Ha, Lee, and
  Davidson]{hafner2019learning}
Danijar Hafner, Timothy Lillicrap, Ian Fischer, Ruben Villegas, David Ha,
  Honglak Lee, and James Davidson.
\newblock Learning latent dynamics for planning from pixels.
\newblock In \emph{International Conference on Machine Learning}, pages
  2555--2565. PMLR, 2019.

\bibitem[Hafner et~al.(2020)Hafner, Lillicrap, Ba, and
  Norouzi]{hafner2020dream}
Danijar Hafner, Timothy Lillicrap, Jimmy Ba, and Mohammad Norouzi.
\newblock Dream to control: Learning behaviors by latent imagination, 2020.

\bibitem[Hendrycks and Gimpel(2020)]{hendrycks2020gaussian}
Dan Hendrycks and Kevin Gimpel.
\newblock Gaussian error linear units (gelus), 2020.

\bibitem[Hennigan et~al.(2020)Hennigan, Cai, Norman, and
  Babuschkin]{haiku2020github}
Tom Hennigan, Trevor Cai, Tamara Norman, and Igor Babuschkin.
\newblock {H}aiku: {S}onnet for {JAX}.
\newblock 2020.
\newblock URL \url{http://github.com/deepmind/dm-haiku}.

\bibitem[Hessel et~al.(2020)Hessel, Budden, Viola, Rosca, Sezener, and
  Hennigan]{optax2020github}
Matteo Hessel, David Budden, Fabio Viola, Mihaela Rosca, Eren Sezener, and Tom
  Hennigan.
\newblock Optax: composable gradient transformation and optimisation, in jax!
\newblock 2020.
\newblock URL \url{http://github.com/deepmind/optax}.

\bibitem[Ibrahim et~al.(2021)Ibrahim, Haworth, Lipani, Aslam, Cheng, and
  Christie]{ibrahim2021variational}
Mohamed~R Ibrahim, James Haworth, Aldo Lipani, Nilufer Aslam, Tao Cheng, and
  Nicola Christie.
\newblock Variational-lstm autoencoder to forecast the spread of coronavirus
  across the globe.
\newblock \emph{PloS one}, 16\penalty0 (1):\penalty0 e0246120, 2021.

\bibitem[Karl et~al.(2017)Karl, Soelch, Bayer, and van~der Smagt]{karl2017deep}
Maximilian Karl, Maximilian Soelch, Justin Bayer, and Patrick van~der Smagt.
\newblock Deep variational bayes filters: Unsupervised learning of state space
  models from raw data, 2017.

\bibitem[Kingma and Welling(2014)]{kingma2014autoencoding}
Diederik~P Kingma and Max Welling.
\newblock Auto-encoding variational bayes, 2014.

\bibitem[Krishnan et~al.(2015)Krishnan, Shalit, and Sontag]{krishnan2015deep}
Rahul~G Krishnan, Uri Shalit, and David Sontag.
\newblock Deep kalman filters.
\newblock \emph{arXiv preprint arXiv:1511.05121}, 2015.

\bibitem[Krishnan et~al.(2016)Krishnan, Shalit, and
  Sontag]{krishnan2016structured}
Rahul~G. Krishnan, Uri Shalit, and David Sontag.
\newblock Structured inference networks for nonlinear state space models, 2016.

\bibitem[Lai et~al.(2021)Lai, Liu, Efros, and Wang]{lai2021video}
Zihang Lai, Sifei Liu, Alexei~A Efros, and Xiaolong Wang.
\newblock Video autoencoder: self-supervised disentanglement of static 3d
  structure and motion.
\newblock In \emph{Proceedings of the IEEE/CVF International Conference on
  Computer Vision}, pages 9730--9740, 2021.

\bibitem[Lee et~al.(2019)Lee, Nagabandi, Abbeel, and Levine]{lee2019stochastic}
Alex~X Lee, Anusha Nagabandi, Pieter Abbeel, and Sergey Levine.
\newblock Stochastic latent actor-critic: Deep reinforcement learning with a
  latent variable model.
\newblock \emph{arXiv preprint arXiv:1907.00953}, 2019.

\bibitem[Pessoa et~al.(2020)Pessoa, Aidos, Tom{\'a}s, and
  Figueiredo]{pessoa2020end}
Jorge Pessoa, Helena Aidos, Pedro Tom{\'a}s, and M{\'a}rio~AT Figueiredo.
\newblock End-to-end learning of video compression using spatio-temporal
  autoencoders.
\newblock In \emph{2020 IEEE Workshop on Signal Processing Systems (SiPS)},
  pages 1--6. IEEE, 2020.

\bibitem[Radford et~al.(2015)Radford, Metz, and
  Chintala]{radford2015unsupervised}
Alec Radford, Luke Metz, and Soumith Chintala.
\newblock Unsupervised representation learning with deep convolutional
  generative adversarial networks.
\newblock \emph{arXiv preprint arXiv:1511.06434}, 2015.

\bibitem[Rezende et~al.(2014)Rezende, Mohamed, and
  Wierstra]{rezende2014stochastic}
Danilo~Jimenez Rezende, Shakir Mohamed, and Daan Wierstra.
\newblock Stochastic backpropagation and approximate inference in deep
  generative models.
\newblock In \emph{International conference on machine learning}, pages
  1278--1286. PMLR, 2014.

\bibitem[Shah et~al.(2017)Shah, Dey, Lovett, and Kapoor]{airsim2017fsr}
Shital Shah, Debadeepta Dey, Chris Lovett, and Ashish Kapoor.
\newblock Airsim: High-fidelity visual and physical simulation for autonomous
  vehicles.
\newblock In \emph{Field and Service Robotics}, 2017.
\newblock URL \url{https://arxiv.org/abs/1705.05065}.

\bibitem[Tanizaki(1996)]{tanizaki1996nonlinear}
Hisashi Tanizaki.
\newblock \emph{Nonlinear filters: estimation and applications}, volume 400.
\newblock Springer Science \& Business Media, 1996.

\bibitem[Terejanu(2009)]{terejanu2009discreteKF}
Gabriel Terejanu.
\newblock Discrete kalman filter tutorial.
\newblock 2009.

\bibitem[Theis et~al.(2017)Theis, Shi, Cunningham, and
  Husz{\'a}r]{theis2017lossy}
Lucas Theis, Wenzhe Shi, Andrew Cunningham, and Ferenc Husz{\'a}r.
\newblock Lossy image compression with compressive autoencoders.
\newblock \emph{arXiv preprint arXiv:1703.00395}, 2017.

\bibitem[van~den Oord et~al.(2017)van~den Oord, Vinyals, and
  Kavukcuoglu]{oord2018neural}
A{\"{a}}ron van~den Oord, Oriol Vinyals, and Koray Kavukcuoglu.
\newblock Neural discrete representation learning.
\newblock \emph{CoRR}, abs/1711.00937, 2017.
\newblock URL \url{http://arxiv.org/abs/1711.00937}.

\bibitem[Watter et~al.(2015)Watter, Springenberg, Boedecker, and
  Riedmiller]{watter2015embed}
Manuel Watter, Jost~Tobias Springenberg, Joschka Boedecker, and Martin
  Riedmiller.
\newblock Embed to control: A locally linear latent dynamics model for control
  from raw images.
\newblock \emph{arXiv preprint arXiv:1506.07365}, 2015.

\bibitem[Yu et~al.(2020)Yu, Dalca, Iglesias, and Sabuncu]{yu2020an}
Evan~M. Yu, Adrian~V. Dalca, Juan~Eugenio Iglesias, and Mert~R. Sabuncu.
\newblock An auto-encoder strategy for adaptive image segmentation.
\newblock In \emph{Medical Imaging with Deep Learning}, 2020.
\newblock URL \url{https://openreview.net/forum?id=aEQCZR3xEm}.

\bibitem[Zhang and Maire(2020)]{zhang2020self}
Xiao Zhang and Michael Maire.
\newblock Self-supervised visual representation learning from hierarchical
  grouping.
\newblock \emph{arXiv preprint arXiv:2012.03044}, 2020.

\end{thebibliography}
